\documentclass{article}
\usepackage[final]{neurips_2024}
\usepackage[T1]{fontenc}    % use 8-bit T1 fonts
\usepackage{hyperref}       % hyperlinks
\usepackage{url}            % simple URL typesetting
\usepackage{booktabs}       % professional-quality tables
\usepackage{amsfonts}       % blackboard math symbols
\usepackage{amsmath,amssymb}
\usepackage{dsfont}
\usepackage{xparse}
\usepackage{nicefrac}       % compact symbols for 1/2, etc.
\usepackage{microtype}      % microtypography
\usepackage{xcolor}         % colors
\usepackage{mathtools}
\usepackage{subcaption}
\usepackage[safe]{tipa} % type phonetic symbols
\usepackage{color, colortbl}
\usepackage{bm}
\usepackage{algorithm}
\usepackage{algpseudocode}
\usepackage{textcomp, gensymb}
\usepackage{wrapfig}
\usepackage[pdftex]{graphicx}
\usepackage{multirow}
\usepackage{enumitem}
\usepackage{amsthm}

\DeclareMathOperator*{\argmax}{argmax}

\title{Robust Fine-tuning of Zero-shot Models via Variance Reduction}

\author{
 \textbf{Beier Zhu} \quad 
 \textbf{Jiequan Cui} \quad
 \textbf{Hanwang Zhang} \\
\small 
Nanyang Technological University\\
\tt \small 
beier002@e.ntu.edu.sg, hanwangzhang@ntu.edu.sg}

\begin{document}

\newcommand{\defeq}{\vcentcolon=}
\newcommand{\eqdef}{=\vcentcolon}

\newtheorem{definition}{Definition}
\newtheorem{theorem}{Theorem}
\newtheorem{assumption}{Assumption}
\newtheorem{lemma}{Lemma}
\newtheorem{proposition}{Proposition}
\newtheorem{corollary}{Corollary}

\def\eg{\emph{e.g.}} 
\def\Eg{\emph{E.g}}
\def\ie{\emph{i.e.}} 
\def\Ie{\emph{I.e}}
\def\cf{\emph{cf} } 
\def\Cf{\emph{Cf}}
\def\etc{\emph{etc}} 
\def\vs{\emph{vs}}
\def\wrt{w.r.t. } 
\def\dof{d.o.f}
\def\iid{i.i.d} 
\def\wolog{w.l.o.g}
\def\etal{\emph{et al}}

\newcommand{\Ours}{VRF }
\newcommand{\ours}{VRF}
\newcommand{\ferm}{f(\cdot;\theta_{\mathsf{ft}})}
\newcommand{\fzs}{f(\cdot;\theta_{\mathsf{zs}})}
\newcommand{\fzsx}{f(\x;\theta_{\mathsf{zs}})}
\newcommand{\fla}{f_{\mathsf{la}}}
\newcommand{\fens}{f_{\mathsf{ens}}}
\newcommand{\Ev}{\Phi_{\mathsf{v}}}
\newcommand{\Et}{\Phi_{\mathsf{t}}}
\newcommand{\softmax}{\mathrm{softmax}}
\newcommand{\x}{\mathbf{x}}
\newcommand{\bt}{\mathbf{t}}
\newcommand{\e}{\mathbf{e}}
\newcommand{\z}{\mathbf{z}}
\newcommand{\q}{\mathbf{q}}
\newcommand{\R}{\mathcal{R}}

\definecolor{tabhighlight}{HTML}{e5e5e5}

\newcommand{\tableCellHeight}{1}
\newcommand{\tabstyle}[1]{
  \setlength{\tabcolsep}{#1}
  \renewcommand{\arraystretch}{\tableCellHeight}
  \centering
  \small
}

\newenvironment{customitemize}[1]{%
    \begin{list}{\labelitemi}{%
        \setlength{\leftmargin}{#1} % Set the left margin/indentation
    }
}{%
    \end{list}
}

\newtheoremstyle{restatedlemma}
  {\topsep}       % Space above
  {\topsep}       % Space below
  {\itshape}      % Body font
  {}              % Indent amount
  {\bfseries}     % Theorem head font
  {.}             % Punctuation after theorem head
  {.5em}          % Space after theorem head
  {\thmname{#1} \thmnumber{#2} (\thmnote{#3})} % Theorem head spec (can be left empty, meaning ‘normal’)

\newtheoremstyle{restatedproposition}
  {\topsep}       % Space above
  {\topsep}       % Space below
  {\itshape}      % Body font
  {}              % Indent amount
  {\bfseries}     % Theorem head font
  {.}             % Punctuation after theorem head
  {.5em}          % Space after theorem head
  {\thmname{#1} \thmnumber{#2} (\thmnote{#3})} % Theorem head spec (can be left empty, meaning ‘normal’)

\theoremstyle{restatedlemma}
\newtheorem*{restatedlemma}{Restated Lemma}

\theoremstyle{restatedproposition}
\newtheorem*{restatedproposition}{Restated Proposition}

\maketitle
\begin{abstract}
When fine-tuning zero-shot models like CLIP, our desideratum is for the fine-tuned model to excel in both in-distribution (ID) and out-of-distribution (OOD).
Recently, ensemble-based models (ESM) have been shown to offer significant robustness improvement, while preserving high ID accuracy. However, our study finds that ESMs do not solve the ID-OOD trade-offs: they achieve peak performance for ID and OOD accuracy at different mixing coefficients. When optimized for OOD accuracy, the ensemble model exhibits a noticeable decline in ID accuracy, and vice versa. In contrast, we propose a sample-wise ensembling technique that can simultaneously attain the best ID and OOD accuracy without the trade-offs. Specifically, we construct a Zero-Shot Failure (ZSF) set containing training samples incorrectly predicted by the zero-shot model. For each test sample, we calculate its distance to the ZSF set and assign a higher weight to the fine-tuned model in the ensemble if the distance is small. We term our method Variance Reduction Fine-tuning (VRF), as it effectively reduces the variance in ensemble predictions, thereby decreasing residual error. On ImageNet and five derived distribution shifts, our \Ours further improves the OOD accuracy by 1.5 - 2.0 pp over the ensemble baselines while maintaining or increasing ID accuracy. \Ours achieves similar large robustness gains (0.9 - 3.1 pp) on other distribution shifts benchmarks. Codes are available in \url{https://github.com/ BeierZhu/VRF}.
\end{abstract}
\section{Introduction}\label{sec:intro}
To ensure the reliability of machine learning systems, it is essential to develop models that can generalize to unseen, out-of-distribution environments. 
Large pre-trained models such as CLIP~\cite{radford2021learning} and ALIGN~\cite{jia2021scaling} have recently shown remarkable robustness against challenging distribution shifts.
However, it is widely acknowledged that these improvements in robustness are most pronounced in the zero-shot setting, while conventional fine-tuning on these models often compromises robustness when compared to zero-shot performance~\cite{wortsman2022robust,kumarfine,pmlr-v180-kumar22a}. This phenomenon is known as the ID-OOD trade-offs, \ie, improving performance on in-distribution (ID) data can sometimes lead to decreased performance on out-of-distribution (OOD) data~\cite{khani2021removing,tripuraneni2020theory}.

In recent years, ensemble-based models (ESMs) have demonstrated significant success in addressing the ID-OOD dilemma~\cite{yong2023spurious,wortsman2022robust,pmlr-v180-kumar22a,zhu2024generalized}. Specifically, denote the input as $\x$, the zero-shot model as $\hat{\mathbb{P}}(y|\x;\theta_\mathsf{zs})$ and the fine-tuned model as $\hat{\mathbb{P}}(y|\x;\theta_\mathsf{ft})$, existing ESMs typically employ the output-space ensemble (OSE)~\cite{pmlr-v180-kumar22a,zhu2024generalized}, which outputs $\hat{\mathbb{P}}(y|\x;\theta_\mathsf{ose})=\alpha\hat{\mathbb{P}}(y|\x;\theta_\mathsf{ft}) + (1-\alpha)\hat{\mathbb{P}}(y|\x;\theta_\mathsf{zs})$, and the weight-space ensemble (WSE)~\cite{wortsman2022robust,yong2023spurious}, which outputs $\hat{\mathbb{P}}(y|\x;\theta_\mathsf{wse})=\hat{\mathbb{P}}(y|\x;\alpha\theta_\mathsf{ft} + (1-\alpha)\theta_\mathsf{zs}),\ \text{where}\ \alpha \in [0,1]$. Compared to fine-tuned models, ESMs offer significant accuracy enhancements under distribution shift, while maintaining high ID accuracy.
% \footnote{In this work, we refer to the reference distribution, on which the fine-tuned models are trained, as the in-distribution (ID), and we refer to the distribution shifts as out-of-distribution (OOD).}

\begin{figure}[t]
    \centering
\includegraphics[width=1.\textwidth]{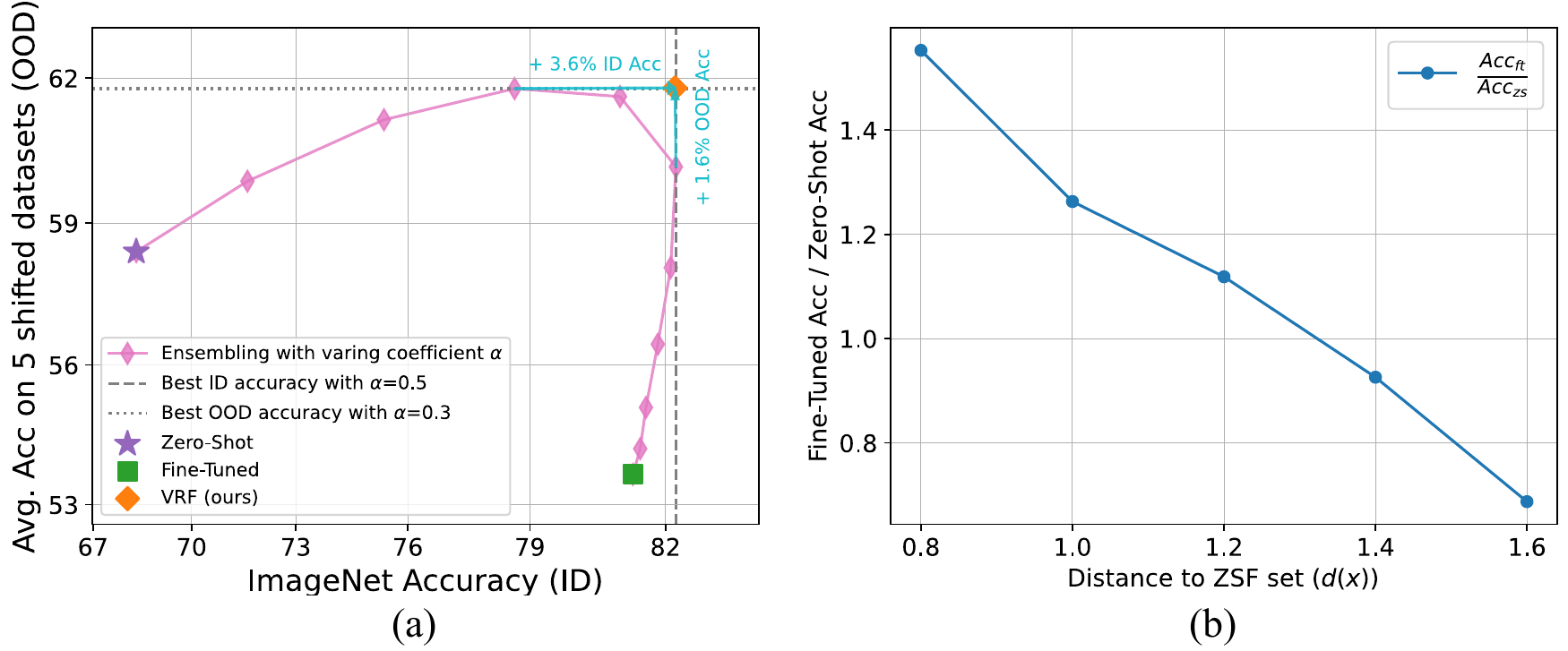}
    \caption{(a) ID-OOD frontier curves for the CLIP ViT-B/16 model on the ID (ImageNet) and OOD (IN-\{V2, R, A, Sketch\} and ObjectNet) datasets by varying the mixing coefficient $\alpha$. The ensemble model achieves its best ID and OOD performance at different $\alpha$ values. Our method \Ours simultaneously attains the best ID and OOD accuracy, outperforming the ensemble by $3.6\%$ on OOD and $1.6\%$ on ID at its optimal performance points.(b) Relationship between the ratio of fine-tuned accuracy to zero-shot accuracy ($\frac{\text{Acc}_\mathsf{ft}}{\text{Acc}_\mathsf{zs}}$) and the distance to the zero-shot failure set ($d(\x)$). 
    $\frac{\text{Acc}_\mathsf{ft}}{\text{Acc}_\mathsf{zs}}$ demonstrates  a monotonic decrease as $d(\x)$ increases.
    }
\label{fig:teaser}
\end{figure}

However, ESM cannot fully address the ID-OOD trade-offs. In Figure~\ref{fig:teaser} (a), by varying the mixing coefficient $\alpha$, we plot the ID-OOD frontier curves ({\color{pink} pink} line) for the CLIP ViT-B/16 model on ImageNet~\cite{deng2009imagenet} (ID) and five derived distribution-shifted datasets  (OOD): ImageNet-V2~\cite{recht2019imagenet}, ImageNet-R~\cite{hendrycks2021many}, ImageNet-A~\cite{hendrycks2021natural}, ImageNet-Sketch~\cite{wang2019learning} and ObjectNet~\cite{barbu2019objectnet}. We find that the ensemble model achieves its optimal ID and OOD performance at different $\alpha$ values: the best ID accuracy is achieved at $\alpha=0.5$ and the best OOD accuracy is obtained at $\alpha=0.3$. When the ensemble model reaches its optimal value for OOD, the performance on ID decreases by $3.6\%$ relative to its peak. Similarly, when the ensemble model is optimized for ID, the performance on OOD decreases by $1.6\%$ relative to its best value -- the ID-OOD trade-offs still persist for ESMs. This raises a natural question:
\begin{center}
\textit{Can ensemble-based models simultaneously attain the best ID and OOD accuracy?}
\end{center}

In this paper, we affirmatively answer this question by proposing a sample-wise ensembling technique, dubbed variance reduction fine-tuning (VRF). This method is motivated by an empirical finding illustrated in Fig~\ref{fig:teaser} (b). For each sample in the training dataset, if the fine-tuned model correctly predicts the label  while the zero-shot model fails, we collect its features representation in the fine-tuned model as the zero-shot failure (ZSF) set. We then measure the distance $d(\x)$ of each test sample $\x$ to the ZSF set. Based on this distance, test samples are grouped into bins, and we compute the ratio of fine-tuned accuracy to zero-shot accuracy: $\frac{\text{Acc}_\mathsf{ft}}{\text{Acc}_\mathsf{zs}}$ for each bin (implementation details are in Section~\ref{app:ft_zs_ratio}). 
Interestingly, we observe that the ratio 
$\frac{\text{Acc}_\mathsf{ft}}{\text{Acc}_\mathsf{zs}}$
  monotonically decreases as $d(\x)$ increases. 
Intuitively, the closer a sample is to the ZSF set, the more likely it is that the zero-shot model makes incorrect predictions, whereas the fine-tuned model is more likely to be accurate, leading to a higher $\frac{\text{Acc}_\mathsf{ft}}{\text{Acc}_\mathsf{zs}}$ ratio.  
Therefore, we use the distance to assign weights to the models: a smaller $d(\x)$ results in a higher weight for the fine-tuned model, and vice versa.

As depicted by the {\color{orange} orange} diamond in Fig.~\ref{fig:teaser} (a), by leveraging the sample-wise weights, our \Ours simultaneously attains the best ID and OOD accuracy. In Section~\ref{sec:exp}, we show that on a variety of different models and tasks, our \Ours approach consistently outperforms the existing fine-tuning and ensembling methods, including linear probing, end-to-end fine-tuning, LP-FT~\cite{kumarfine}, OSE and WSE~\cite{wortsman2022robust}. 
In specific, on ImageNet and five derived distribution shifts, our VRF further improves
the OOD accuracy by 1.5 - 2.0 pp over the ensemble baselines while maintaining
or increasing ID accuracy.
Furthermore, in Section~\ref{sec:justify}, we justify our approach by demonstrating that it effectively minimizes the variance of the ensemble models, resulting in reduced residual error.

\section{Related Work}
\noindent\textbf{Mitigating ID-OOD trade-offs.}
Improving performance on in-distribution data can sometimes lead to a decrease in performance on out-of-distribution data, and vice versa. This phenomenon is known as the ID-OOD trade-offs. 
% Previous research has employed self-training on large amounts of unlabeled data to counteract these trade-offs~\cite{xie2020n}. 
Xie et al.~\cite{xie2020n} leverage auxiliary information as outputs of auxiliary tasks to pre-train a model to reduce OOD error. Khani and Liang~\cite{khani2021removing} show that
self-training on large amounts of unlabeled data can mitigate such trade-offs by removing spurious features. Tripuraneni et al.~\cite{tripuraneni2020theory} tackle this problem by learning representations that are robust across diverse tasks. However, these methods usually necessitate additional unlabeled data or auxiliary information. In contrast, our VRF is a straightforward variation of fine-tuning that does not require any extra data.

\noindent\textbf{Robust fine-tuning of zero-shot models.}
Vision-language models like CLIP~\cite{radford2021learning} have demonstrated outstanding improvements in robustness. It is commonly acknowledged that conventional fine-tuning methods often compromise robustness when compared to zero-shot performance. 
Therefore, enhancing downstream robustness has been the focus of subsequent works~\cite{kumarfine,wortsman2022robust,goyal2022finetune,nam2024lipsum,han2024anchor,zhu2023debiased}.
Kumar et al.~\cite{kumarfine} show that a two-process of linear probing followed by full fine-tuning can alleviate feature distortion, leading to stronger OOD performance without sacrificing ID accuracy. Wortsman et al.~\cite{wortsman2022robust} propose a method of weight interpolation between the zero-shot and the fine-tuned models to improve both ID and OOD accuracy. 
% Zhu et al.~\cite{zhu2023debiased} leverage dynamic knowledge distillation to achieve a balanced performance between accuracy and robustness. 
Goyal et al.~\cite{goyal2022finetune} demonstrate that mimicking the contrastive pre-training objectives to fine-tune the zero-shot models outperforms tuning via the traditional supervised cross-entropy loss. 
However, the ID-OOD trade-offs are still observed with these methods. In contrast, our method \Ours can simultaneously achieve the best ID and OOD accuracy.
\section{Methods}\label{sec:methods}

\subsection{Set Up}\label{sec:setup}
\noindent\textbf{Task:} Consider a classification setting where the goal is to map an instance $\x \in \mathcal{X}$ to a label $y \in \mathcal{Y}=[K]$. We are provided with a zero-shot model $\fzs$, a downstream dataset $\mathcal{D}=\{\x_i,y_i\}_{i=1}^N$, and a fine-tuned model $\ferm$  which is trained on $\mathcal{D}$. Below, we outline the implementation of the zero-shot and fine-tuned models:

\begin{itemize}[leftmargin=8mm]
    \item \textbf{Zero-shot models} (ZS): We investigate CLIP models~\cite{radford2021learning} as our zero-shot models. CLIP models are pre-trained using image-text pairs $\{(\x_1,\bt_1),...,(\x_B,\bt_B)\}$ from the Internet. The objective of the CLIP models is to train a visual encoder $\Phi_\textsf{v}$ and a text encoder $\Phi_\textsf{t}$ such that the cosine similarity $<\Phi_\textsf{v}(\x_i),\Phi_\textsf{t}(\bt_i)>$ is maximized relative to unmatched pairs. CLIP models perform zero-shot inference for $K$ classes by matching $\x$ with potential class names $\{c_1,...,c_K\}$. Concretely, by extending the class name $\{c_k\}$ to a prompt ``$\bt_k=$a photo of a $\{c_k\}$'', the zero-shot model outputs the score (logit) for class $k$ as $\fzsx_k=<\Phi_\textsf{v}(\x),\Phi_\textsf{t}(\bt_k)>$. The predicted probabilities can be calculated using the softmax function, \ie, $\hat{\mathbb{P}}(y|\x;\theta_\mathsf{zs})=\text{softmax}(\fzsx)_y$.
    The model outputs the label as $\text{pred}(f(\x;\theta_\mathsf{zs}))=\argmax_i f(\x;\theta_\mathsf{zs})_i  $
    \item \textbf{Linear classifiers} (LC): We learn a linear classifier on top of the visual embedding $\Phi_\mathsf{v}(\x)$ while freezing the visual encoder $\Phi_\mathsf{v}$. The parameters of the linear classifier are optimized to minimize the cross-entropy loss on $\mathcal{D}$.
    \item \textbf{End-to-end fine-tuning} (E2E-FT): We update both the linear classifier and the visual encoder by minimizing the cross-entropy loss on $\mathcal{D}$.
    \item \textbf{Linear probing then full fine-tuning}~\cite{kumarfine} (LP-FT): We employ a two-phase fine-tuning approach: initially training a linear classifier, followed by full fine-tuning starting from the solution derived from training the linear classifier.
    \item \textbf{Output-space ensemble} (OSE): We perform linear interpolation of the outputs between a zero-shot model and a fine-tuned model (\eg, E2E-FT or LP-FT):
        \begin{equation}
            \hat{\mathbb{P}}(y|\x;\theta_\mathsf{ose})=\alpha\hat{\mathbb{P}}(y|\x;\theta_\mathsf{ft}) + (1-\alpha)\hat{\mathbb{P}}(y|\x;\theta_\mathsf{zs}),\ \text{where}\ \alpha \in [0,1]
        \end{equation}
    \item \textbf{Weight-space ensemble}~\cite{wortsman2022robust} (WSE). We combine the weights through linear interpolation between a zero-shot model and a fine-tuned model: 
        \begin{equation}
            \hat{\mathbb{P}}(y|\x;\theta_\mathsf{wse})=\hat{\mathbb{P}}(y|\x;\alpha\theta_\mathsf{ft} + (1-\alpha)\theta_\mathsf{zs}),\ \text{where}\ \alpha \in [0,1]
        \end{equation}
\end{itemize}

\subsection{Variance Reduction Fine-tuning}
We now present our proposed method, \ours, which consists of three steps. First, before the inference stage, we collect the Zero-Shot Failure (ZSF) set. Second, for a given test sample, we calculate its distance to the ZSF set. Third, we assign weights to combine predictions from the zero-shot and fine-tuned models based on this distance.

\noindent\textbf{Step 1 (Identification).}
For each $\x_i$ in the training dataset $\mathcal{D}$, if the fine-tuned model correctly predicts the label while the zero-shot model fails, we collect its feature representation $\mathbf{v}_i = \Phi_\mathsf{v}(\x_i; \theta_\mathsf{ft})$ from the fine-tuned model to form the zero-shot failure set $\mathcal{V}$. Specifically, $\mathcal{V}$ is defined as:
\begin{equation}\label{eq:zsf}
  \mathcal{V} = \{\mathbf{v}_i \ \text{s.t.} \ y_i = \text{pred}(f_\mathsf{ft}(\x_i)) \ \text{and} \ y_i \neq \text{pred}(f_\mathsf{zs}(\x_i))\}.
\end{equation}
Here, $f_\mathsf{zs}(\cdot)$ and $f_\mathsf{ft}(\cdot)$ are used to denote $\fzs$ and $\ferm$, respectively, for simplicity.

\noindent\textbf{Step 2 (Distance Calculation).} The key empirical observation underpinning \Ours is that in the vicinity of the ZSF set, a test sample typically exhibits lower zero-shot accuracy ($\text{Acc}_\mathsf{zs}$) and higher fine-tuned accuracy ($\text{Acc}_\mathsf{ft}$). Consequently, the $\frac{\text{Acc}_\mathsf{ft}}{\text{Acc}_\mathsf{zs}}$ ratio demonstrates a monotonic decrease as the distance from the sample to the ZSF set increases.
In this paper, we adopt non-parametric density estimation using nearest neighbors~\cite{sun2022out} to measure the distance of a test sample to the dataset $\mathcal{V}$.
Specifically, during inference, we derive the feature representation $\mathbf{v}$ of a test sample $\x$, and compute the $\ell_2$ distances $\|\mathbf{v}-\mathbf{v}_i\|_2$ \wrt $\mathbf{v}_i \in \mathcal{V}$.
We reorder $\mathcal{V}$ according to the increasing $\ell_2$ distance and denote the ordered set in sequence as $\mathcal{V}'=(\mathbf{v}_{(1)},\mathbf{v}_{(2)},...,\mathbf{v}_{(|\mathcal{V}|)})$.
The distance of $\x$ to $\mathcal{V}$ is defined as the $\ell_2$ distance to the $k$-th nearest neighbor ($k$-NN), \ie,
\begin{equation}\label{eq:knn-distance}
    d(\x;\mathcal{V},k)=\|\mathbf{v}-\mathbf{v}_{(k)}\|_2.
\end{equation}
If there is no ambiguity, we use $d(\x)$ to denote $ d(\x;\mathcal{V},k)$ for readability. Since the features in CLIP models are $\ell_2$ normalized, $d(\x)$ are bounded between $[0,2]$.

\begin{wrapfigure}{r}{0.4\textwidth}
    \centering
\includegraphics[width=0.4\textwidth]{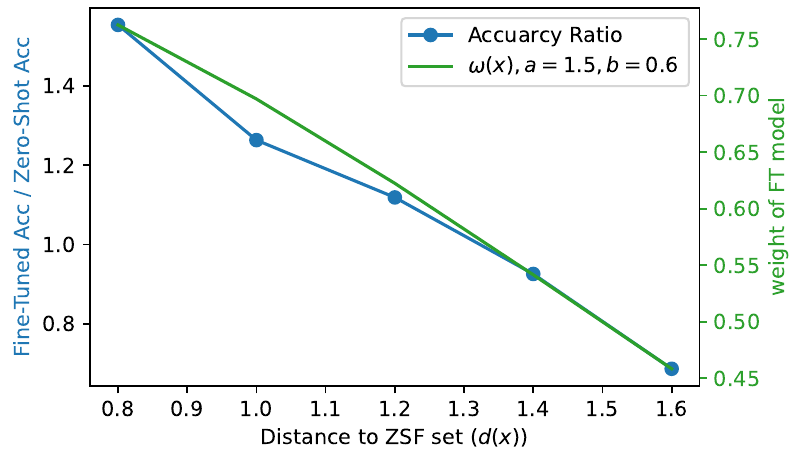}
    \caption{Relationship between $\frac{\text{Acc}_\mathsf{ft}}{\text{Acc}_\mathsf{zs}}$ and the weight $\omega(\x)$.}
    \label{fig:ft_zc_w}
    \vspace{2mm}
\end{wrapfigure}
\noindent\textbf{Step 3 (Sample-Wise Ensembling).} We implement sample-wise out-space ensembling in the form:
\begin{equation}\label{eq:vrf}
    \hat{\mathbb{P}}_\mathsf{vrf}(y|\x) = \omega(\x)\cdot \hat{\mathbb{P}}_\mathsf{ft}(y|\x) + (1-\omega(\x))\cdot\hat{\mathbb{P}}_\mathsf{zs}(y|\x),
\end{equation}
where $\omega(\x) \in (0,1)$. We use the distance to ZSF set $d(\x)$ to determine the weight $\omega$. As shown by the {\color{blue} blue} line in Fig~\ref{fig:ft_zc_w}, a smaller value of $d(\x)$ corresponds to a larger $\frac{\text{Acc}_\mathsf{ft}}{\text{Acc}_\mathsf{zs}}$ ratio, and vice versa. Therefore, we set the weight $\omega$ to be inversely proportional to $d(\x)$. Given that $\omega$ is bounded between $0$ and $1$, we employ a sigmoid function $\sigma(\cdot)$ as:
\begin{equation}\label{eq:weight}
    \omega(\x) = \sigma(-(d(\x) - a)/b),
\end{equation}
where $a,b > 0$ are two hyper-parameters sweeped using accuracy on ID validation set. 
We visualize the weight curve in {\color{green} green} on Fig~\ref{fig:ft_zc_w}, setting $a=1.5$ and $b=0.6$. We summarize the whole process in Algorithm~\ref{algo:1}.  
\begin{algorithm}[t]
\caption{Variation Reduction Fine-tuning}
\begin{algorithmic}[1]
\State \textbf{Given}: Training dataset $\mathcal{D}$, a zero-shot model $f_\mathsf{zs}$ and a fine-tuned model $f_\mathsf{ft}$.

\State Build zero-shot failure set $\mathcal{V}$ using Eq.~\eqref{eq:zsf}.\Comment{Step 1: Identification}
\State \textbf{Inference Stage:}
\State Given a test sample $\x$, compute its feature representation $\mathbf{v}$, zero-shot prediction $\hat{\mathbb{P}}_\mathsf{zs}(y|\x)$ and fine-tuned model prediction $\hat{\mathbb{P}}_\mathsf{ft}(y|\x)$.
\State Compute the $k$-NN distance to $\mathcal{V}$ as $d(\x)$ using Eq.~\eqref{eq:knn-distance}. \Comment{Step 2: Distance Calculation}
\State Compute the weight $\omega(\x)$ using Eq.~\eqref{eq:weight}.
\State Return $\hat{\mathbb{P}}_\mathsf{vrf}(y|\x)$ using Eq.~\eqref{eq:vrf} \Comment{Step 3: Sample-Wise Ensembling}
\end{algorithmic}\label{algo:1}
\end{algorithm}
\section{Justification}\label{sec:justify}
We now prove that our \Ours can effectively reduce the variance of the combined model, resulting in lower errors compared to ensembling using a constant mixing coefficient. 

\subsection{Background}
The outputs of a well trained classifier are expected to approximate the \textit{a posterior} class distribution. 
Apart from the irreducible error (Bayes error), the residual error of a classifier can be broken down into bias and variance components. In specific, for a test sample $\x$, the probability output of a classifier parameterized by $\theta$ can be expressed as:
\begin{equation}
    \hat{\mathbb{P}}(y|\x;\theta) = \mathbb{P}(y|\x) + \underbrace{\beta_y + \eta_y(\x)}_{\text{residual error for } \x},
\end{equation}
where $\mathbb{P}(y|\x)$ denotes the true \textit{a posterior} distribution, $\beta_y$ is the label bias of $\hat{\mathbb{P}}(y|\x;\theta)$ which is independent to the input $\x$, and $\eta_y(\x)$ is related to the given input $\x$. In this study, we primarily attribute the residual error to the variance term (\ie, $\beta_y=0$), as the label bias problem in foundation models has been effectively addressed by Zhu et al.~\cite{zhu2024generalized}. Tumer et al.~\cite{tumer1996analysis} have proven that the expected residual error $E$ is given by: 
\begin{equation}\label{eq:var-error}
    E=\frac{\mathbb{V}[\eta_y(\x)]}{s},
\end{equation}
where $s$ is a constant factor related to the derivative of the true \textit{a posterior} distribution and is independent of the trained model, and $\mathbb{V}[\eta_y(\x)]$ is the variance.

\subsection{Variance Reduction Fine-tuning Leads to Lower Residual Error}
Let us shift our focus to the effects of combining the zero-shot and fine-tuned models. 
Let $g_\mathsf{zs}(\cdot)$ and $g_\mathsf{ft}(\cdot)$ be two functions that produce weights for ensembling the models. Subject to the constraint that $g_\mathsf{zs}(\x) + g_\mathsf{ft}(\x)=1$,
the residual error of the combined classifier is obtained by:
\begin{equation}
    \hat{\mathbb{P}}_\mathsf{vrf}(y|\x)=g_\mathsf{zs}(\x) \hat{\mathbb{P}}_\mathsf{zs}(y|\x) + g_\mathsf{ft}(\x) \hat{\mathbb{P}}_\mathsf{ft}(y|\x)  =\mathbb{P}(y|\x) + \underbrace{g_\mathsf{zs}(\x)\cdot \eta_\mathsf{zs}(\x) + g_\mathsf{ft}(\x)\cdot \eta_\mathsf{ft}(\x)}_{\eta_\mathsf{vrf}(\x)} ,
\end{equation}
where we omit the subscript $y$ of $\eta$ for readability. The variance of $\eta_\mathsf{vrf}(\x)$ can be expressed as:
\begin{equation}\label{eq:simplevar}
    \mathbb{V}[\eta_\mathsf{vrf}(\x)] = g_\mathsf{zs}(\x)^2\cdot \mathbb{V}[\eta_\mathsf{zs}(\x)] + g_\mathsf{ft}(\x)^2\cdot \mathbb{V}[\eta_\mathsf{ft}(\x)].
\end{equation}
Here, we assume the residual errors are independent following the assumption of the previous studies of CLIP fine-tuning~\cite{pmlr-v180-kumar22a,zhu2024generalized}. We further explore the case of correlated residual errors in Section~\ref{app:correlated}.  
According to Eq.~\eqref{eq:var-error}, the reduction in variance can be readily translated into a reduction in error rates. To obtain the smallest variance $\mathbb{V}[\eta_\mathsf{vrf}(\x)]$, we minimize Eq.~\eqref{eq:simplevar} using Lagrange multiplier to enforce the constraint that $g_\mathsf{zs}(\x) + g_\mathsf{ft}(\x)=1$, and obtain the optimal weight function $g_\mathsf{ft}$ as:
\begin{equation}
     g_\mathsf{ft}(\x)=\frac{\mathbb{V}[\eta_\mathsf{zs}(\x)]}{\mathbb{V}[\eta_\mathsf{zs}(\x)] + \mathbb{V}[\eta_\mathsf{ft}(\x)]}=\frac{E_\mathsf{zs}}{E_\mathsf{zs}+E_\mathsf{ft}}=(1+\frac{E_\mathsf{ft}}{E_\mathsf{zs}})^{-1} \propto \frac{\text{Acc}_\mathsf{ft}}{\text{Acc}_\mathsf{zs}}
\end{equation}
Since $\frac{\text{Acc}_\mathsf{ft}}{\text{Acc}_\mathsf{zs}} \propto d(\x)^{-1}$ (a smaller distance $d(\x)$ is associated with a larger $\frac{\text{Acc}_\mathsf{ft}}{\text{Acc}_\mathsf{zs}}$ as shown in Fig.~\ref{fig:ft_zc_w}), we design the weighting function $g_\mathsf{ft}(\x)=\omega(\x)  \propto d(\x)^{-1} $ as in Eq.~\eqref{eq:weight}.  
\section{Experiments}\label{sec:exp}
\subsection{Experimental Setup}
\noindent\textbf{Datasets with distribution shifts.} 
We provide the results for ImageNet~\cite{deng2009imagenet} and its five derived distribution shifts: (1) ImageNet-V2 (IN-V2)~\cite{recht2019imagenet}: Test images sampled after a decade of the original ImageNet. (2) ImageNet-R (IN-R)~\cite{hendrycks2021many}: Contains renditions (\eg,  art, cartoons, graffiti). (3) ImageNet Sketch (IN-Sketch)~\cite{wang2019learning}: Consists of sketches rather than natural photos. (4) ImageNet-A (IN-A)~\cite{hendrycks2021natural}: Collects real-world images that are misclassified by ResNet models. (5) ObjectNet~\cite{barbu2019objectnet}, a test set featuring objects with diverse backgrounds, rotations, and imaging viewpoints.   
We extend our analysis to include a standard distribution shift benchmark~\cite{kumarfine,pmlr-v180-kumar22a,french2018self}: CIFAR-10 $\rightarrow$ STL-10, where the ID is CIFAR-10~\cite{krizhevsky2009learning}, and the OOD is STL-10~\cite{coates2011analysis}. We removed the ``monkey'' class from STL-10, as it does not exist in CIFAR-10. In addition, we also consider subpopulation shifts, where the ID data contains a few sub-categories, and the OOD data comprises different sub-categories within the same parent category. Following~\cite{kumarfine,pmlr-v180-kumar22a}, we adopt Entity30 dataset~\cite{Santurkar2020BREEDSBF},  which aims to categorize images into one of 30 entity categories, such as ``vehicle'' and ``insect''.

\noindent\textbf{Baselines.}
We adopt two models: CLIP ViT-B/32 and a larger ViT-B/16 from OpenAI~\cite{radford2021learning}. The default model used in ablation studies is the CLIP ViT-B/16.
In addition to the zero-shot models, we compare our approach against five standard methods for adapting pre-trained models: (1) linear classifier~\cite{radford2021learning}, (2) E2E-FT, (3) LP-FT~\cite{kumarfine}, (4) OSE, and (5) WSE~\cite{wortsman2022robust}. The descriptions of these methods have been included in Section~\ref{sec:setup}.

\noindent\textbf{Implementation details.} 
When fine-tuning E2E-FT models, we adhere to Wortsman et al.~\cite{wortsman2022robust}, employing the default PyTorch AdamW optimizer for 10 epochs with weight decay of 0.1 and a cosine-annealing learning rate schedule with 500 warm-up steps. Unless specified, we use a learning rate of $3\times 10^{-5}$, gradient clipping at norm 1. When fine-tuning LP-FT, we first adopt the settings of Wortsman et al.~\cite{wortsman2022robust} to train the linear classifier, then full fine-tune the models at a learning rate of $1\times 10^{-5}$. 
For efficiently performing $k$-NN search, we use Faiss library~\cite{johnson2019billion}. Denote the size of the ZSF set is $|\mathcal{V}|$, we scale the $k$ according to a percentage $p\%$ of the sample set, where $k=\text{floor}(p\%\cdot |\mathcal{V}|)$. In this paper, $p$ is set to $0.1\%$, a value consistent with the default setting proposed by Sun et al.~\cite{sun2022out}. 
Note that all the hyperparameters, \eg, $\alpha, a, b$, are searched using the accuracy on the in-distribution (ID) validation set. Derived distribution shift datasets are \textit{only for evaluation and not for hyperparameter sweeps.} See Appendix~\ref{app:exp-detail} for the details of experimental details.
\begin{table}
\caption{Accuracy of various methods on ImageNet and derived distribution shifts for CLIP ViT-B/32.}
\label{tab:imagenet_vit32}
  \centering
    \tabstyle{4pt}
    \begin{tabular}{l c|ccccc|c}
    \toprule
      \multirow{2}{*}{Method} & \multirow{2}{*}{IN} & \multicolumn{5}{c|}{Distribution shifts} & Avg \\  
     &  & IN-V2 & IN-Sketch & IN-A & IN-R & ObjectNet & shifts \\
    \midrule
    Zero-shot~\cite{radford2021learning} &  63.3 &	55.9 &	42.3 &	31.5 &	69.3 & 43.5 & 48.5 \\
    Linear classifier~\cite{radford2021learning} & 75.4	&63.4	&38.8	&26.1&	58.7 & 41.5 & 45.7 \\ \midrule
    E2E-FT ~\cite{wortsman2022robust} & 76.2 &	64.2 & 	38.7 &	21.0 & 57.1&  40.1 & 44.2 \\
    \quad + Weight-space ensemble ~\cite{wortsman2022robust} & 77.9 & 67.2 & 45.1 &	28.8 & 66.4& 45.1 & 50.5\\
    \quad + Output-space ensemble & 77.3 & 66.0 & 44.2 & 27.1 & 68.4 & 44.4 & 50.0 \\
    \quad + \Ours (ours) & 77.6 & 66.7 & 47.0  & 29.2  & 70.9& 46.3 & 52.0\\
    \rowcolor{tabhighlight}
    \quad $\Delta$ & +0.3 & +0.7 & +2.8 & +2.1 & +2.5 & +1.9 & +2.0 \\
    \midrule
    LP-FT~\cite{kumarfine} & 76.9 &	64.8 &	39.9 &	25.7 &	69.9& 42.6 & 48.6 \\
    \quad + Weight-space Ensemble ~\cite{wortsman2022robust} & 78.0 & 	67.0 & 44.8 & 31.2 & 65.8 & 46.1 &	51.0  \\
    \quad + Output-space Ensemble & 77.8 &	66.3 &	44.0 & 29.5 & 66.2 & 45.5 &	50.3  \\ 
    \quad + VRF (ours) & 77.8 & 66.7 & 	46.1 & 31.0 & 	70.0 & 46.3 & 51.8	 \\
    \rowcolor{tabhighlight}
    \quad $\Delta$ & +0.0 & +0.4 & +2.1 & +1.5 & +3.8 & +0.8 & +1.5 \\
    \bottomrule
    \end{tabular}
\end{table}

\begin{table}
 \caption{Accuracy of various methods on ImageNet and derived distribution shifts for CLIP ViT-B/16.}
 \label{tab:imagenet_vit16}
  \centering
    \tabstyle{4pt}
    \begin{tabular}{l c|ccccc|c}
    \toprule
      \multirow{2}{*}{Method} & \multirow{2}{*}{IN} & \multicolumn{5}{c|}{Distribution shifts} & Avg \\  
     &  & IN-V2 & IN-Sketch & IN-A & IN-R & ObjectNet & shifts \\
    \midrule
    Zero-shot~\cite{radford2021learning} &  68.3 & 61.9 & 48.3 & 50.1 &	77.6 &54.2 & 58.4\\
    Linear classifier~\cite{radford2021learning} & 79.3 &	69.1 & 44.8 & 44.3 & 66.7 & 51.1 & 55.2 \\ \midrule
    E2E-FT ~\cite{wortsman2022robust} & 81.3 & 70.6 & 45.1 & 36.6 & 65.6 & 50.5 & 53.7\\
    \quad + Weight-space ensemble ~\cite{wortsman2022robust} & 82.5 & 73.1 & 51.6 & 47.6 &  75.1 & 55.7 & 60.6\\
    \quad + Output-space ensemble &82.2 & 	72.0 &	50.6 &	46.8 &	76.7 & 54.9 & 60.2 \\
    \quad + \Ours (ours) & 82.3 &  72.1 & 52.9 &	48.4  & 78.7 &56.4 & 61.8  \\ 
    \rowcolor{tabhighlight}
    \quad $\Delta$ & +0.1 & +0.1 & +2.3 & +1.6 & +2.0 & +1.5 & +1.6 \\ \midrule
    LP-FT~\cite{kumarfine} & 81.5 &	70.7 &	46.7 &	41.4 & 66.4 & 52.4 &  55.5\\
    \quad + Weight-space ensemble ~\cite{wortsman2022robust} & 82.4 &	73.0 &	51.5 &	50.6 &	74.2 &56.6&  61.2\\
    \quad + Output-space ensemble & 82.1 &	72.3 &	50.9 &	50.9 &	74.9 &	55.7 & 60.9 \\
    \quad + \Ours (ours) & 82.1 &	72.3 &	52.9 & 51.2 &	78.8 & 57.2 &  62.4\\
    \rowcolor{tabhighlight}
    \quad $\Delta$ & +0.0 & +0.0 & +2.0 & +0.3 & +3.9 & +1.5 & +1.5 \\
    \bottomrule
    \end{tabular}
\end{table}

\subsection{Results}

\noindent\textbf{ImageNet and its five shifted distribution results.} In Table~\ref{tab:imagenet_vit32} and \ref{tab:imagenet_vit16}, we report the ID-OOD accuracies of fine-tuning baselines for CLIP ViT-32 and CLIP ViT-16 models, respectively. For OSE and WSE, we choose the mixing coefficient $\alpha$ with the highest ID validation accuracy. To enhance clarity in the results, we denote the improvement over OSE as $\Delta$ in Tables~\ref{tab:imagenet_vit32} and \ref{tab:imagenet_vit16}. We observe that our \Ours boosts the accuracy of fine-tuned models, including ensembling baseline models, across five ImageNet distribution shifted datasets, while maintaining or improving the ImageNet in-distribution performance. For instance, in Table~\ref{tab:imagenet_vit32}, when ensembling with the E2E-FT model, our \Ours outperforms the OSE model by $2.0\%$ on distribution shifts while increasing the ID accuracy by $0.3\%$. Compared to WSE models, our \Ours achieves a delta of $1.2\%$ on distribution shifts, while maintaining ID performance within $0.2\%$, as shown in E2E-FT part of Table~\ref{tab:imagenet_vit16}.  

\setlength\intextsep{0pt}
\begin{table}[t]
\centering
\caption{Accuracy of various methods on CIFAR-10 $\rightarrow$ STL-10 and Entity-30.}
\tabstyle{4pt}
\begin{subtable}[t]{.45\textwidth}
    \centering
    \begin{tabular}{l|cc|cc}
    \toprule
    \multirow{2}{*}{Method}         &
    \multicolumn{2}{c|}{CIFAR $\rightarrow$ STL} & \multicolumn{2}{c}{Entity-30} \\
    & ID  & OOD & ID & OOD \\ \midrule
    Zero-shot~\cite{radford2021learning} & 88.3 & 97.1  &65.2 &	66.5 \\
    Linear classifier & 95.0 &	96.6& 93.3 &	68.1 \\ \midrule
    E2E-FT~\cite{wortsman2022robust} & 97.9 &	93.5&94.4 &	65.1   \\
    \quad + WSE~\cite{wortsman2022robust} & 98.2 &	95.7 & 94.6 &	68.8 \\
    \quad + OSE & 97.9 & 95.9& 94.4 &	66.4 \\
    \quad + \Ours (ours) & 97.8 &	97.3 & 94.5	& 69.5 \\
    \rowcolor{tabhighlight}
    \quad $\Delta$ & -0.1 & +1.4 & +0.1 & +3.1 \\ \midrule
    LP-FT~\cite{kumarfine} & 97.9 &	95.0& 94.6 & 67.7 \\
    \quad + WSE~\cite{wortsman2022robust} & 98.1 &	96.4 & 94.8 & 68.8\\
    \quad + OSE & 98.1 & 96.4 & 94.7 & 68.5\\
    \quad + \Ours (ours) & 98.1	& 97.5 & 94.8 & 70.1 \\
    \rowcolor{tabhighlight}
    \quad $\Delta$ & +0.0 & +1.1& +0.1 & +1.6  \\
   \bottomrule
    \end{tabular}
    \caption{CLIP ViT-B/32}
\end{subtable}
\quad\quad
\begin{subtable}[t]{.45\textwidth}
    \centering
    \begin{tabular}{l|cc|cc}
    \toprule
    \multirow{2}{*}{Method}         &
    \multicolumn{2}{c|}{CIFAR $\rightarrow$ STL} & \multicolumn{2}{c}{Entity-30} \\
    & ID  & OOD & ID & OOD \\ \midrule 
    Zero-shot~\cite{radford2021learning} & 90.1	& 98.4& 68.3 &	68.2 \\
    Linear classifier & 95.8	& 97.7& 95.3&69.6  \\ \midrule
    E2E-FT~\cite{wortsman2022robust} & 98.6	& 96.1& 96.9 & 68.2\\
    \quad + WSE~\cite{wortsman2022robust} & 98.7 & 97.8&97.2	& 71.9 \\
    \quad + OSE & 98.6 & 96.6 & 97.0  &	71.5 \\
    \quad + \Ours (ours) & 98.6	& 98.4& 97.0 &	72.7 \\
    \rowcolor{tabhighlight}
    \quad $\Delta$ & +0.0 & +1.8 & +0.0 & +1.2\\ \midrule
    LP-FT~\cite{kumarfine} & 98.5 & 96.3 & 96.9 & 68.8\\
    \quad + WSE~\cite{wortsman2022robust} & 98.7 & 97.9&  97.3 & 72.1 \\
    \quad + OSE & 98.6 & 97.7 & 97.2 & 71.8\\
    \quad + \Ours (ours) & 98.6	 & 98.6 &  97.4 & 72.9\\
    \rowcolor{tabhighlight}
    \quad $\Delta$ & +0.0 & +0.9 & +0.2 &+1.1\\
   \bottomrule
    \end{tabular}
    \caption{CLIP ViT-B/16}
\end{subtable}
\label{tab:CIFAR10results}
\vspace{-4mm}
\end{table}
\noindent\textbf{CIFAR-10 $\rightarrow$ STL-10 and Entity-30 results.} We report the accuracy of various methods in Table~\ref{tab:CIFAR10results} (a,b). We note that fine-tuning baselines can enhance the accuracy on CIFAR-10 compared to the zero-shot models. However, this improvement comes at the expense of reduced accuracy on STL-10. For instance, E2E-FT leads to a decrease of approximately $3.6\%$ in STL-10 accuracy, as shown in Table~\ref{tab:CIFAR10results}(a). Previous ensemble methods can mitigate the degradation to some extent, but the STL-10 performance still lags behind the zero-shot performance, \eg, In Table~\ref{tab:CIFAR10results}(b), the accuracy of E2E-FT + WSE is $97.8\%$ whereas the zero-shot performance is $98.4\%$.
In contrast, our \Ours simultaneously improves accuracy on both CIFAR-10 and STL-10. Similarly, for Entity-30, our \Ours can further improvement the OOD performance when compared to WSE and OSE methods. 
\begin{figure}[t]
    \centering
\includegraphics[width=1.0\textwidth]{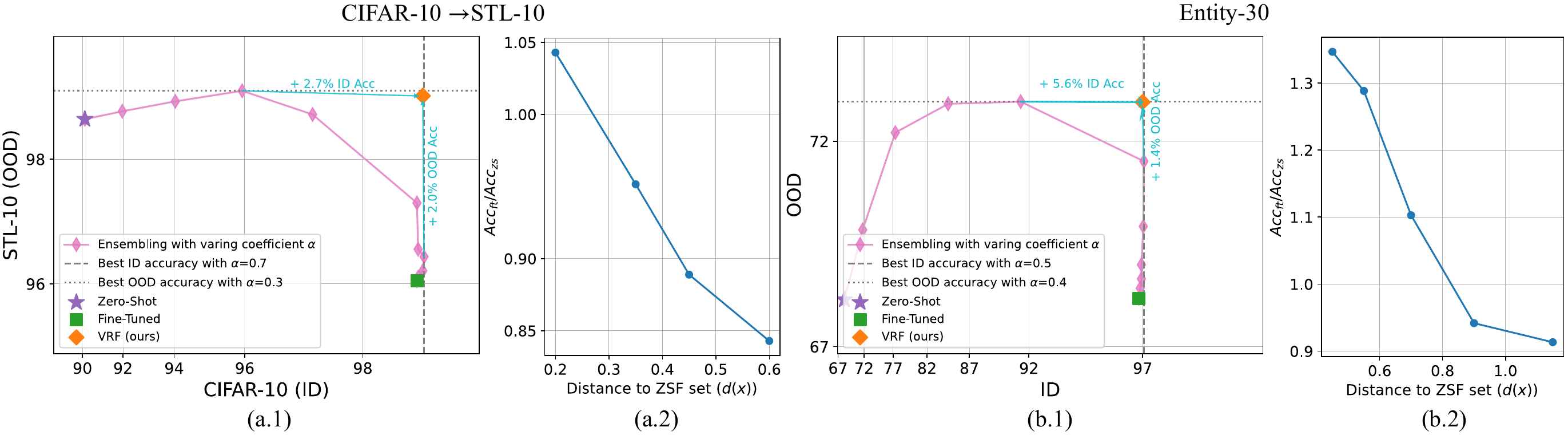}
    \caption{ ID-OOD frontier curves by varying the mixing coefficient $\alpha$ and $\frac{\text{Acc}_\mathsf{ft}}{\text{Acc}_\mathsf{zs}}$ curves for the CLIP ViT-B/16 . (a) CIFAR-10 (ID) and STL-10 (OOD) results. (b) Entity-30 results.
    }
\label{fig:id-ood-ft-zs-acc}
\end{figure}

In addition, we plot the ID-OOD frontier curves in Figure~\ref{fig:id-ood-ft-zs-acc} (a.1\&b.1), respectively. Similar to the results on ImageNet (Figure~\ref{fig:teaser}(a)), the ensemble model achieves its best ID and OOD performances at different $\alpha$ values. For instance, on the CIFAR-10 benchmark, when the ensemble model attains its optimal ID value at $\alpha=0.7$, the OOD performance decreases by $2.0\%$ relative to its peak. Conversely, when the optimal OOD value is reached at $\alpha=0.3$, the performance on ID diminishes by $2.7\%$ from its best. In contrast, our VRF simultaneously attains the ID and OOD performance. 

We also analyze the relation between the ratio $\frac{\text{Acc}_\mathsf{ft}}{\text{Acc}_\mathsf{zs}}$ and $d(\x)$ in Figure~\ref{fig:id-ood-ft-zs-acc} (a.2\&b.2). Consistent with the findings from ImageNet (Figure~\ref{fig:teaser} (b)), we observe that the ratio decreases as $d(\x)$ increases, which further supports our design of assigning a higher weight to fine-tuned models if $d(\x)$ is smaller.

\begin{table}
\centering
\tabstyle{4pt}
\caption{Results of VRF for linear-probed models using CLIP ViT-B/16 models.}
\begin{tabular}{l|cc|cc|cc}
    \toprule
    \multirow{2}{*}{Method} &
    \multicolumn{2}{c|}{ImageNet}         &
    \multicolumn{2}{c|}{CIFAR-10} & \multicolumn{2}{c}{Entity-30} \\
    & ID  & OOD & ID & OOD & ID & OOD \\ \midrule
    Zero-shot classifier~\cite{radford2021learning} &68.3 & 58.4 & 90.1 & 98.4  & 68.3 & 68.2	 \\
    Linear classifier & 79.3 & 55.2 & 95.8  &	97.7 & 95.3 & 69.6 \\ \midrule
    WSE/OSE & 79.9 & 57.8 & 95.8 & 97.7 & 95.5  &	70.5  \\
    \Ours (ours) & 79.8 & 58.5 & 95.8  & 98.4 & 95.4	& 71.4  \\
   \bottomrule
\end{tabular}\label{tab:vrf_linear}
\end{table}
\subsection{Further Analysis and Ablation Studies}\label{sec:abla}
\noindent\textbf{VRF for linear-probed models.} A drawback of the proposed method is its doubled inference and storage cost compared to WSE and other single-model robust fine-tuning methods. To address concerns regarding space-time complexity, we apply our VRF method to linear-probed models and present the results in  Table~\ref{tab:vrf_linear}. We also compare with output-space ensembling, since the model is linear, it is equivalent to weight-space ensembling. We also compare it with output-space ensembling, which, given the linear nature of the model, is equivalent to weight-space ensembling. Consistent with the conclusions drawn from fully fine-tuned models, our \Ours method further improves OOD performance while maintaining comparable ID performance to OSE/WSE ensembling.

\begin{wrapfigure}{r}{0.4\textwidth}
    \centering
\includegraphics[width=0.4\textwidth]{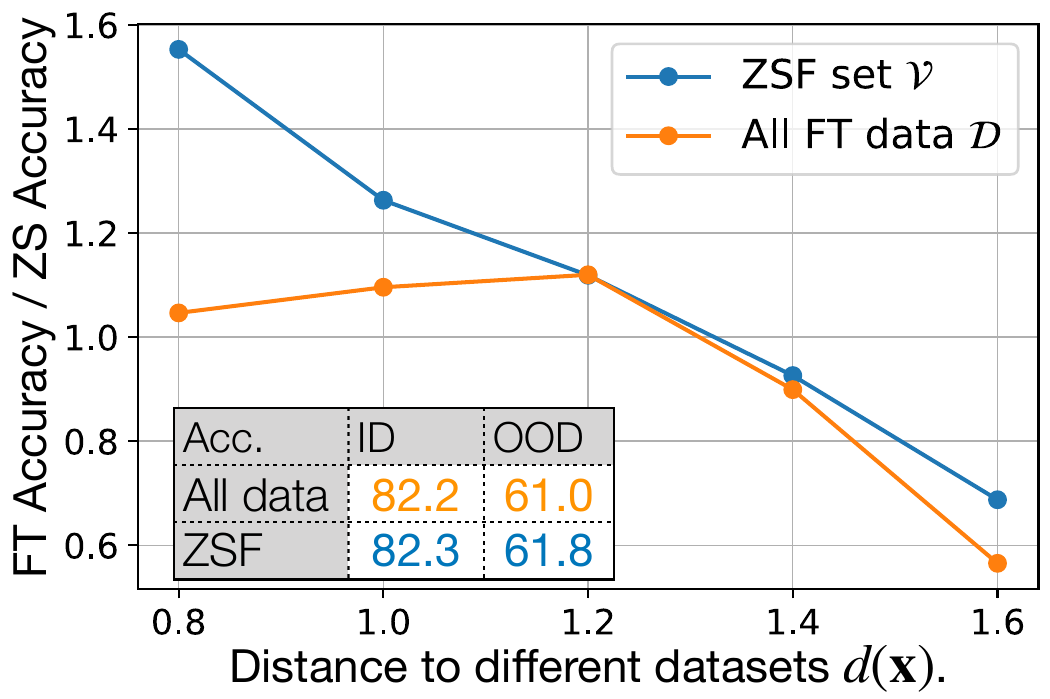}
    \caption{ZSF set $\mathcal{V}$ vs. all data $\mathcal{D}$}
    \label{fig:zsf_all}
    % \vspace{-2mm}
\end{wrapfigure}
\noindent\textbf{Using ZSF set $\mathcal{V}$ or entire training set $\mathcal{D}$?}
In Step 1 of our \ours, we define the zero-shot failure set $\mathcal{V}$ and use it to compute distances. We aim to find out whether using the entire training set $\mathcal{D}$ offers comparable performance. In Figure~\ref{fig:zsf_all}, we plot the $\frac{\text{Acc}_\mathsf{ft}}{\text{Acc}_\mathsf{zs}}$ curves and report both ID and OOD accuracy using the two sets. We observe that the ratio curve using $\mathcal{D}$ does not exhibit a monotonic trend with $d(\x)$: it initially increases and then decreases as $d(\x)$ increases. 
Furthermore, the ratio $\frac{\text{Acc}_\mathsf{ft}}{\text{Acc}_\mathsf{zs}}$ using $\mathcal{D}$ is less informative when $d(\x)$ is smaller than 1.2, as the curve relatively remains flat.  
As the zero-shot models can accurately predict a large proportion of the ID data (recall that the zero-shot accuracy is $68.3\%$), a smaller distance to entire training set $\mathcal{D}$ does not reliably indicate whether the fine-tuned model can make more accurate predictions. In comparison, our ZSF set contains only the samples where zero-shot models fail but fine-tuned models succeed. When a sample is close to $\mathcal{V}$, it is more likely that the accuracy ratio will be high. Consequently, the performances using $\mathcal{D}$ are clearly outperformed by those using $\mathcal{V}$.

\begin{wraptable}{r}{4cm} 
\tabstyle{6pt}
\caption{Selective prediction using OOD detector.}
\begin{tabular}{l cc}
    \toprule 
    Method & ID & OOD \\
    \midrule
    % Zero-shot &  68.3 & 58.4\\
    % E2E-FT & 81.3 & 53.7\\ \midrule
    MSP~\cite{hendrycks2016baseline}     & 81.5 & 57.3\\
    Energy~\cite{liu2020energy}  & 81.0 & 57.6 \\
    MD~\cite{lee2018simple}      & 81.0 & 57.7 \\
    kNN~\cite{sun2022out}     & 80.8 & 58.4  \\
    RMD~\cite{ren2021simple}     & 81.1 & 58.4 \\ \midrule
    \Ours (ours) & 82.3 & 61.8 \\
    \bottomrule
\end{tabular}
\vspace{-12mm}
\label{tab:ood_detect}
\end{wraptable}
\noindent\textbf{Comparison with selective prediction using OOD detector.} A simple approach to address the ID-OOD trade-offs is to use an OOD detector for selective prediction. The OOD detector is a binary classifier $G_\lambda(\cdot)$ to decide whether a sample is ID or OOD based on a threshold $\lambda$. 
For a test sample, predictions are made with the fine-tuned model if classified as ID, and with the zero-shot model otherwise:
\begin{align}
    f_\mathsf{sp}(\x)&=\begin{cases} 
   f_\mathsf{ft}(\x) & \text{if } G_\lambda(\x) =  \text{ID} \\
    f_\mathsf{zs}(\x) & \text{if } G_\lambda(\x) =  \text{OOD},  
    \end{cases} \\
    G_\lambda(\x) &=\begin{cases} 
   \text{ID} & \text{if } S(\x) \geq \lambda \\
    \text{OOD} & \text{if } S(\x) < \lambda,  
    \end{cases}, 
\end{align}
    
where instances with higher scores $S(\x)$ are classified as ID and vice versa.
 $\lambda$ is typically chosen to achieve achieve a $95\%$ true positive rate for ID samples. We report the results of several implementations of $S(\x)$ in Table~\ref{tab:ood_detect} (Details are in Section~\ref{app:sp}). 
We note that selective prediction achieves comparable ID performance to E2E-FT models and similar OOD performance to zero-shot models. However, its accuracy still falls significantly short of our VRF.
This is because traditional OOD detectors are designed for scenarios where the OOD data have a completely disjoint label space from the ID data, \ie, $\mathcal{Y}_\mathsf{OOD} \cap \mathcal{Y}_\mathsf{ID}=\emptyset$. However, in our setup, the zero-shot models show predictive power on ID data, and the fine-tuned models are effective on OOD data. Making binary selections may overlook the complementary knowledge from the other model.
Instead, our weight function $\omega(\x)$ intelligently selects the contribution of each model based on the distance to the ZSF set. Another reason why our method outperforms selective prediction is the effective use of the ZSF set, as illustrated in Figure~\ref{fig:zsf_all}. Directly using all ID data as traditional OOD detectors (e.g., kNN and MD) leads to a weak correlation between the accuracy ratio and the distance $d(\x)$ (or score $S(\x)$)

\begin{figure}[t]
    \centering
\includegraphics[width=0.95\textwidth]{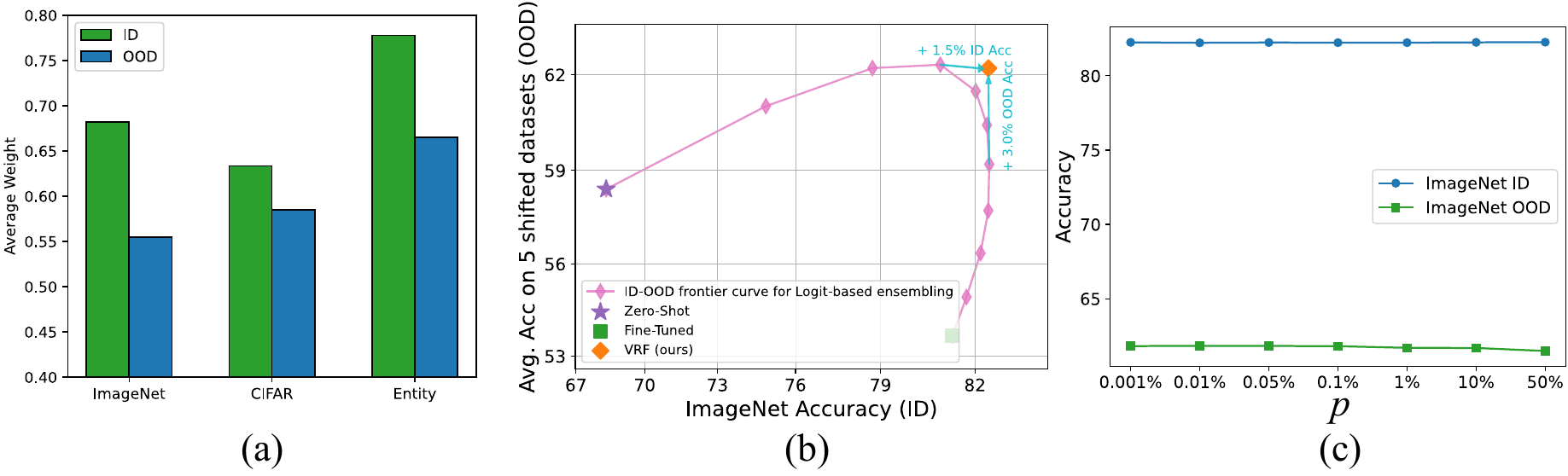}
    \caption{(a) Averaged weight $\mathbb{E}_{\x}[\omega(\x)]$ on different datasets. (b) \Ours based on logit-space ensembling. (c) Comparison with the effect of different $k$ in the $k$-NN distance.
    % All experiments are conducted on CLIP ViT-B/16.
    }
    \vspace{-4mm}
\label{fig:avg-weight}
\end{figure}

\noindent\textbf{Examination of the averaged weight for ID and OOD test sets.} Figure~\ref{fig:avg-weight}(a) shows the average weight ($\mathbb{E}_{\x}[\omega(\x)]$) of the E2E-FT model in ensembling for both ID and OOD test sets. As expected, higher average weights are observed in the ID test set, as the fine-tuned models excel in such domain.

\noindent\textbf{Logits-based ensembling.}  In this paper, we implement OSE by linearly interpolating the probabilities of the two models. Another common strategy for ensembling, known as Logits-Space Ensembling (LSE), involves interpolating in the logits space: $f(\x;\theta_\mathsf{lse})=\alpha f(\x;\theta_\mathsf{ft}) + (1-\alpha) f(\x;\theta_\mathsf{zs})$. We aim to investigate whether our \Ours can enhance the robustness of LSE without compromising the ID accuracy. The results depicted in Figure~\ref{fig:avg-weight}(b) confirm that our \Ours can indeed generalize to LSE.

\begin{figure}[t]
    \centering
\includegraphics[width=0.95\textwidth]{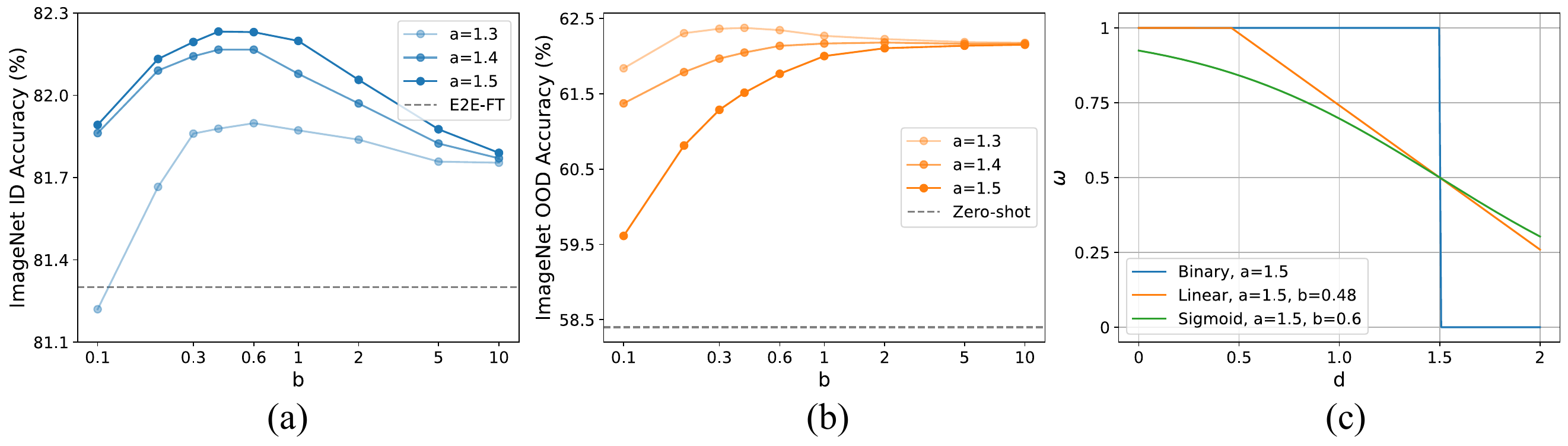}
    \caption{(a) Effect of $a$ and $b$ on ImageNet ID accuracy. (b) Effect of $a$ and $b$ on ImageNet OOD accuracy. (c) Other designs of $\omega(\x)$, hyper-parameters are searched on validation set.}
\label{fig:ab-ablation}
\vspace{-6mm}
\end{figure}

\noindent\textbf{Effect of $k$ in $k$-NN distance.}
In Figure~\ref{fig:avg-weight}(c), to compute $k=\text{floor}(p\cdot|\mathcal{V}|)$, we vary $p$ across the range $\{0.0001\%, 0.01\%, 0.05\%, 0.1\%, 10\%, 50\%\}$. We note two observations: (1) Varying $k$ slightly affect the ID performance: the fluctuations are less than $0.1\%$. (2) The OOD accuracy declines as $p$ increases, but the degradation is very slight for relative small values of $p$ 
(\eg, when $p<0.01\%$, the decline is smaller than $0.2\%$). In our implementation, we use the $k$-th nearest sample instead of the nearest one to reduce the potential impact of label noise. 
If the nearest sample is mislabeled, the distance may be unreliable. The $k$-th sample, being in a higher-density region, offers more stable distance estimates with lower variance, as it lies between the $(k-1)$-th and $(k+1)$-th samples. This makes the measure more robust to outliers. Additionally, prior research~\cite{sun2022out} shows that using the $k$-th nearest distance improves density estimation, which we adopt here.

% \begin{wrapfigure}{r}{0.3\textwidth}
%  \caption{Accuracy of designs of $\omega$ on ImageNet.}
%  \label{tab:design-of-w}
%   \centering
%     \tabstyle{6pt}
%     \begin{tabular}{l cc}
%     \toprule 
%     Design of $\omega$ & ID & OOD \\
%     \midrule
%     % Zero-shot &  68.3 & 58.4\\
%     % E2E-FT & 81.3 & 53.7\\
%     Binary  & 81.3 & 55.0\\
%     Linear  & 82.3 & 61.7 \\
%     Sigmoid & 82.3 & 61.8  \\ 
%     \bottomrule
%     \end{tabular}
% \end{wrapfigure}

\begin{minipage}{0.3\linewidth} % 调整宽度为页面宽度的一半
\makeatletter\def\@captype{table}
\centering
\tabstyle{4pt}
\caption{Accuracy of designs of $\omega$ on ImageNet.}
 \label{tab:design-of-w}
  \centering
    \tabstyle{6pt}
    \begin{tabular}{l cc}
    \toprule 
    Design of $\omega$ & ID & OOD \\
    \midrule
    % Zero-shot &  68.3 & 58.4\\
    % E2E-FT & 81.3 & 53.7\\
    Binary  & 81.3 & 58.4\\
    Linear  & 82.3 & 61.7 \\
    Sigmoid & 82.3 & 61.8  \\ 
    \bottomrule
    \end{tabular}
\end{minipage}
\hfill
\begin{minipage}{0.65\linewidth}
% \begin{figure}[t]
    \centering
    \includegraphics[width=1\textwidth]{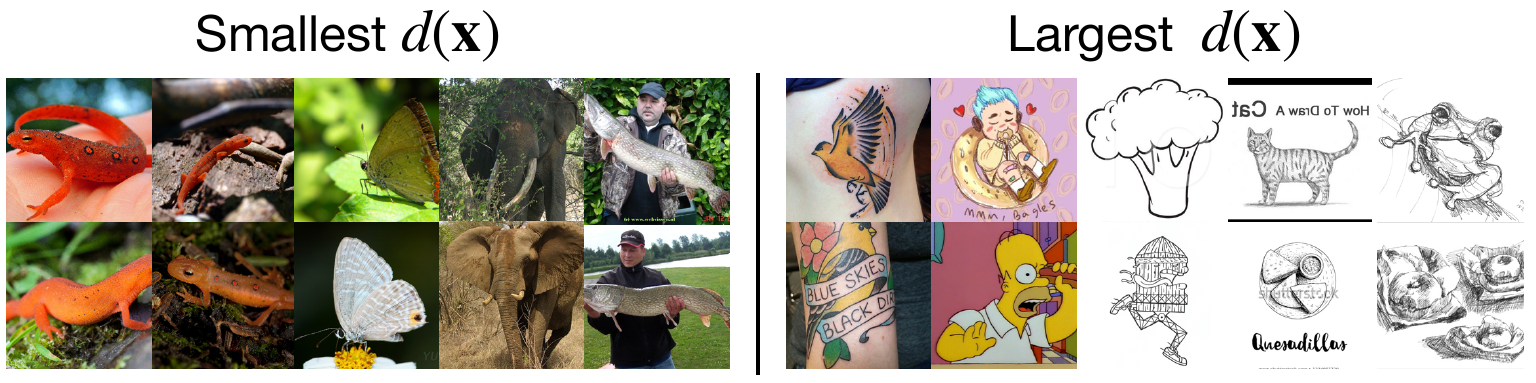}
    \captionof{figure}{Visualization the samples with the smallest/largest $d(\x)$.}
    \label{fig:vis}
% \end{figure}
\end{minipage}

\begin{wrapfigure}{r}{0.35\textwidth}
    \centering
\includegraphics[width=0.35\textwidth]{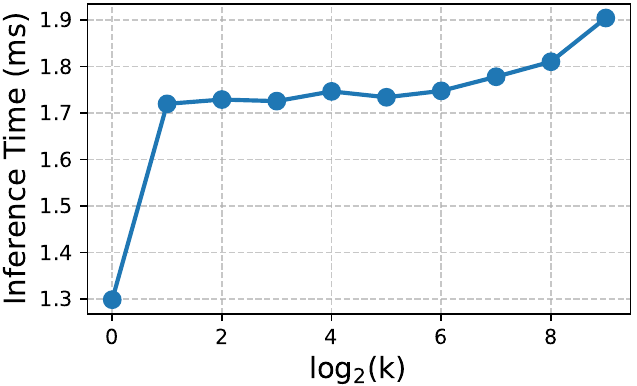}
    \caption{Inference speed (per-image) using different $k$.}
    \label{fig:inference_time}
\end{wrapfigure}
\noindent\textbf{Inference speed of computing $k$-nearest neighbor distance.} 
Thanks to the Faiss library~\cite{johnson2019billion}, the $k$-NN search can be efficiently implemented. 
When evaluated on ImageNet benchmarks using CLIP ViT/B-16 features, the inference speed is approximately $1.8$ milliseconds per-image, which does not significantly improve the inference time. In Figure~\ref{fig:inference_time}, we further present the per-image inference speed of the $k$-nearest neighbor distance computation for various $k$ values. The inference speed is less than $2$ ms when $k < 512$.

\noindent\textbf{Effect of $a$ and $b$ in $\omega$.} We demonstrate the effect of $a$ and $b$ in Figure~\ref{fig:ab-ablation} (a\&b).
% In this paper, we search $a$ and $b$ by find the optimal performance on ID validation set: $a^*=1.5, b^*=0.6$, resulting in $82.3\%$ ID accuracy and $61.8\%$ OOD accuracy. 
We highlight three trends: (1) ID performance peaks at $b\approx0.6$ across different values of $a$. (2) OOD performance often improves as $b$ increases across different values of $a$. (3) When $b$ is sufficiently large, \eg, $b>10$ for ID and $b>2$ for OOD, $a$ has marginal effect on the performance of ID and OOD. In Appendix~\ref{sec:vrf_tradeoffs}, we further plot the trade-offs when tuning $a$ and $b$.

\noindent\textbf{Other designs of $\omega(\x)$.} We further explore alternative designs of $\omega(\x)$ beyond the sigmoid format in Eq.~\eqref{eq:weight}: 
\begin{itemize}
    \item Binary weight: $\omega_\mathsf{binary}(\x)= \mathds{1}[d(\x)<a]$, where $a \in [0,2]$ and $\mathds{1}[\cdot]$ is the indicator function.
    \item Linear weight: $\omega_\mathsf{linear}(\x)=\text{clamp}_{[0,1]}(-b\cdot (d(\x)-a))$, where $a\in [0,2]$, $b > 0$ and $\text{clamp}_{[0,1]}(\cdot)$ rectifies the weight within $[0,1]$.
\end{itemize}
 We report the results on ImageNet in Table~\ref{tab:design-of-w} and plot the weight curves with the value of hyper-parameters in Figure~\ref{fig:ab-ablation}(c). We find that the Linear and the Sigmoid weights show comparable performance and assign similar values of $\omega$ around $d=1.5$. 

\noindent\textbf{Visualization of samples $\x$ according to $d(\x)$.} In Figure~\ref{fig:vis}, we randomly sample testing images with the top-100 smallest $d(\x)$ values 
in the range $[0.40, 0.62]$ 
and the top-100 largest $d(\x)$ values in the range $ \in [1.59, 1.63]$. Interesting, we observe that: (1) Samples with the smallest  $d(\x)$  predominantly consist of fine-grained species, \eg, ``Triturus vulgaris'', ``eft'' and ``lycaenid'', where the fine-tuned models possess domain-specific knowledge, which is often lacking in the zero-shot models. (2) Images with the largest $d(\x)$ exhibit styles different from those of the fine-tuning samples, including tattoos, cartoons, and sketches, contrasting with the photos typically seen in fine-tuning. Zero-shot models are more skilled in non-real photo styles compared to fine-tuned models. 
\section{Impact, limitations  and conclusion}\label{sec:conclusion}
\noindent\textbf{Impact.} 
Zero-shot models inherit the weaknesses from pre-training data to downstream tasks, such as noisy and malicious samples. Our \Ours  might propagate the negative impact.

\noindent\textbf{Limitations.} Our approach is built on the premise that zero-shot models posses predictive capabilities for downstream tasks. However, if the pre-training knowledge significantly differs from the downstream tasks, our algorithm might fail, which is also an open problem in transfer learning.
In addition, the proposed method doubles inference cost compared to WSE and other fine-tuning methods, as it runs both the zero-shot and fine-tuned models. However, this overhead can be mitigated by parallel execution.

\noindent\textbf{Conclusion.}
Inspired by the ID-OOD trade-offs in ensemble-based fine-tuning, we propose VRF to simultaneously optimize the best ID and OOD accuracy. By leveraging the distance to the ZSF set, we assign sample-wise weights to the two models. Despite its simplicity, our \Ours demonstrates strong empirical performance, offering a promising technique for solving ID-OOD trade-offs. 
% We anticipate that our ensemble approach will encourage broader adoption in the pursuit of robustness.

\newpage
\section*{Acknowledgments}
This research is supported by the National Research Foundation, Singapore
under its AI Singapore Programme (AISG Award No: AISG2-PhD-2021-01-002). 
\bibliographystyle{plain}
\bibliography{cite}

\appendix
\newpage

\tableofcontents

\section{Licenses}\label{app:licenses}
All the datasets we considered are publicly available, we list their licences and URLs as follows:

\begin{itemize}[leftmargin=8mm]
    \item \textbf{CIFAR-10}~\cite{krizhevsky2009learning}: MIT License, \url{https://www.cs.toronto.edu/~kriz/cifar.html}.
    \item \textbf{STL-10}~\cite{coates2011analysis}: Non-commercial, \url{https://cs.stanford.edu/~acoates/stl10/}.
    \item \textbf{Entity-30}~\cite{Santurkar2020BREEDSBF}: Non-commercial, \url{https://github.com/MadryLab/BREEDS-Benchmarks}.
    \item \textbf{ImageNet}~\cite{deng2009imagenet}: Non-commercial, \url{http://image-net.org}.
    \item \textbf{IN-V2}~\cite{recht2019imagenet}: MIT License, \url{https://github.com/modestyachts/ImageNetV2}.
    \item \textbf{IN-R}~\cite{hendrycks2021many}: MIT License, \url{https://github.com/hendrycks/imagenet-r}.
    \item \textbf{IN-Sketch}~\cite{wang2019learning}: MIT License, \url{https://github.com/HaohanWang/ImageNet-Sketch}.
    \item \textbf{IN-A}~\cite{hendrycks2021natural}: MIT License, \url{https://github.com/hendrycks/natural-adv-examples}.
    \item \textbf{ObjectNet}~\cite{barbu2019objectnet}: Creative Commons Attribution 4.0,  \url{https://objectnet.dev}.
\end{itemize}
\section{Analysis in the Presence of Correlated Errors}\label{app:correlated}
Our assumption of independent residual errors is based on the previous studies~\cite{wortsman2022robust} (Section 5), where an empirical phenomena is observed: the zero-shot and the fine-tuned models produce diverse predictions.
In general (\ie, the fine-tuned models are initialized from the zero-shot models), we cannot assume that the errors in the zero-shot and fine-tuned models are totally uncorrelated. Let $\mathbb{C}[\eta_\mathsf{zs}(\x),\eta_\mathsf{ft}(\x)]$ be the covariance between $\eta_\mathsf{zs}(\x)$ and $\eta_\mathsf{ft}(\x)$, the variance of $\eta_\mathsf{vrf}(\x)$ can be expressed as:
\begin{equation}\label{eq:fullvar}
    \mathbb{V}[\eta_\mathsf{vrf}(\x)]=g_\mathsf{zs}(\x)^2\cdot \mathbb{V}[\eta_\mathsf{zs}(\x)] + g_\mathsf{ft}(\x)^2\cdot \mathbb{V}[\eta_\mathsf{ft}(\x)] + 2 \cdot g_\mathsf{zs}(\x)\cdot g_\mathsf{ft}(\x)\cdot \mathbb{C}[\eta_\mathsf{zs}(\x),\eta_\mathsf{ft}(\x)].
\end{equation}
Maintaining that $g_\mathsf{zs}(\x)+g_\mathsf{ft}(\x)=1$, the optimal weight $g_\mathsf{ft}^*(\x)$ to minimize Eq.~\eqref{eq:fullvar} becomes:
\begin{equation}\label{eq:cov_error}
    g_\mathsf{ft}^*(\x)=(1+\frac{\mathbb{V}[\eta_\mathsf{ft}(\x)]-\mathbb{C}[\eta_\mathsf{zs}(\x),\eta_\mathsf{ft}(\x)]}{\mathbb{V}[\eta_\mathsf{zs}(\x)]-\mathbb{C}[\eta_\mathsf{zs}(\x),\eta_\mathsf{ft}(\x)]})^{-1}
\end{equation}
\begin{wrapfigure}{r}{0.5\textwidth}
    \centering
\includegraphics[width=0.5\textwidth]{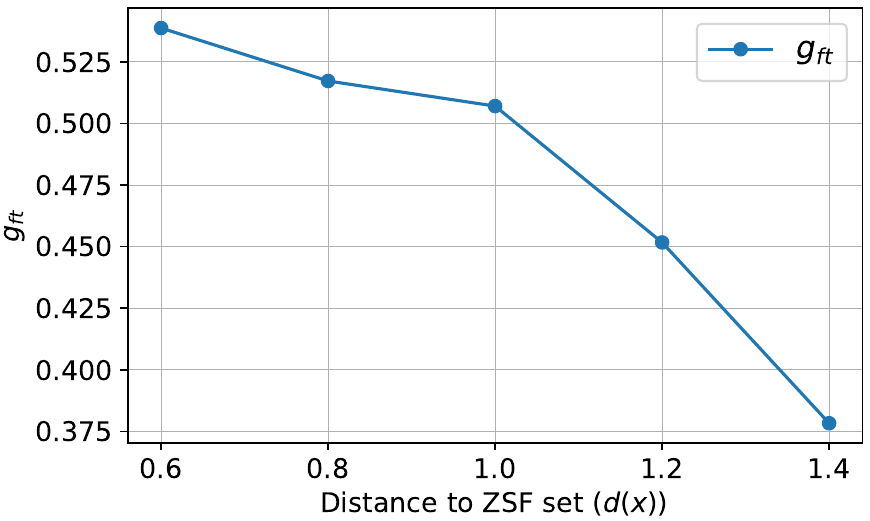}
    \caption{Relationship between $g_\mathsf{ft}(\x)$ and $d(\x)$.}
    \label{fig:cov}
    \vspace{1mm}
\end{wrapfigure}
% Denote the variance of $\eta_\mathsf{ft}(\x)$ as $\sigma_\mathsf{ft}^2(\x)$, the variance of $\eta_\mathsf{zs}(\x)$ as $\sigma_\mathsf{zs}^2(\x)$, and the Pearson correlation coefficient between  $\eta_\mathsf{ft}(\x)$ and $\eta_\mathsf{zs}(\x)$ as $\rho(\x)$, we can rewrite Eq.~\eqref{eq:cov_error} as:
% \begin{equation}
%     g_\mathsf{ft}(\x)=(1+\frac{\sigma_\mathsf{ft}^2(\x)-\rho(\x)\sigma_\mathsf{ft}(\x)\sigma_\mathsf{zs}(\x)}{\sigma_\mathsf{zs}^2(\x)-\rho(\x)\sigma_\mathsf{ft}(\x)\sigma_\mathsf{zs}(\x)})^{-1}
% \end{equation}
Recall that we are interested in using the distance to ZSF set, \ie, $d(\x)$, to surrogate $g^*_\mathsf{ft}(\x)$. To understand the relationship between $d(\x)$ and $g^*_\mathsf{ft}(\x)$, we first group test samples in ImageNet and its five distribution shifted datasets into bins based on the value of $d(\x)$. We then compute the averaged $g^*_\mathsf{ft}(\x)$ for each bin and plot the relationship in Figure~\ref{fig:cov}. In specific, we use temperature scaling~\cite{pmlr-v180-kumar22a} to calibrate the zero-shot and fine-tuned models over the validation set. Afterwards, we define the true distribution $\mathbb{P}(y|\x)$ as a one-hot vector, where the value of 1 corresponds to the true label for a given input $\x$. We then calculate $\eta_\mathsf{ft}(\x)=\hat{\mathbb{P}}(y|\x;\theta_\mathsf{ft})-\mathbb{P}(y|\x)$ and $\eta_\mathsf{zs}(\x)=\hat{\mathbb{P}}(y|\x;\theta_\mathsf{zs})-\mathbb{P}(y|\x)$. Finally, we compute the average $g^*_\mathsf{ft}(\x)$ for each bin as shown in Figure~\ref{fig:cov}. Interestingly, we observe the similar trend in Figure~\ref{fig:teaser}~(b): the weight for fine-tuned models decreases as $d(\x)$ increases. This phenomena indicates that our weighting function $\omega(\x)$ derived under the assumption of independent errors is also valid in the presence of correlated errors.
\section{Additional Experimental Details and Results}

\subsection{Additional Experimental Details}\label{app:exp-detail}
\begin{table}
\centering
\tabstyle{4pt}
\caption{Hyper-parameters $a$ and $b$ for different backbones and datasets.}
\begin{tabular}{l|cc|cc|cc}
    \toprule
    \multirow{2}{*}{Backbone} &
    \multicolumn{2}{c|}{ImageNet}         &
    \multicolumn{2}{c|}{CIFAR-10} & \multicolumn{2}{c}{Entity-30} \\
    & $a$  & $b$ & $a$ & $b$ & $a$ & $b$ \\ \midrule
    CLIP ViT-B/32 & 1.5 & 0.6 & 0.3 &  0.3 & 1.1 & 0.6 	 \\
    CLIP ViT-B/16 & 1.5 & 0.5 & 0.3  & 0.3 & 1.1 & 0.6  \\ 
   \bottomrule
\end{tabular}\label{tab:hyper-parameters}
\end{table}
For CLIP ViT-32 based E2E-FT and LP-FT models, we use a batch size of 512. For CLIP ViT-16 based E2E-FT, we directly download the fine-tuned models from Wortsman et al.~\cite{wortsman2022robust}\footnote{\url{https://drive.google.com/drive/folders/1f56kjpRKPiNSaUxNDtETEDRkbDkZnpCQ}}. The batch size for training CLIP ViT-16 based LP-FT models is set to 384, which is the largest batch size that fits into 2 A6000 GPUs. When performing linear probing, we use a batch of 512 and the initial learning rate of 0.1 for all experiments. The mixing coefficient $\alpha$ for OSE and WSE are searched over $[0,0.1,0.2,...,0.9,1.0]$. The values of $a$ and $b$ for our \Ours are reported in Table~\ref{tab:hyper-parameters}.

\begin{figure}[t]
    \centering
\begin{minipage}[t]{0.32\textwidth}
        \centering
\includegraphics[width=\linewidth]{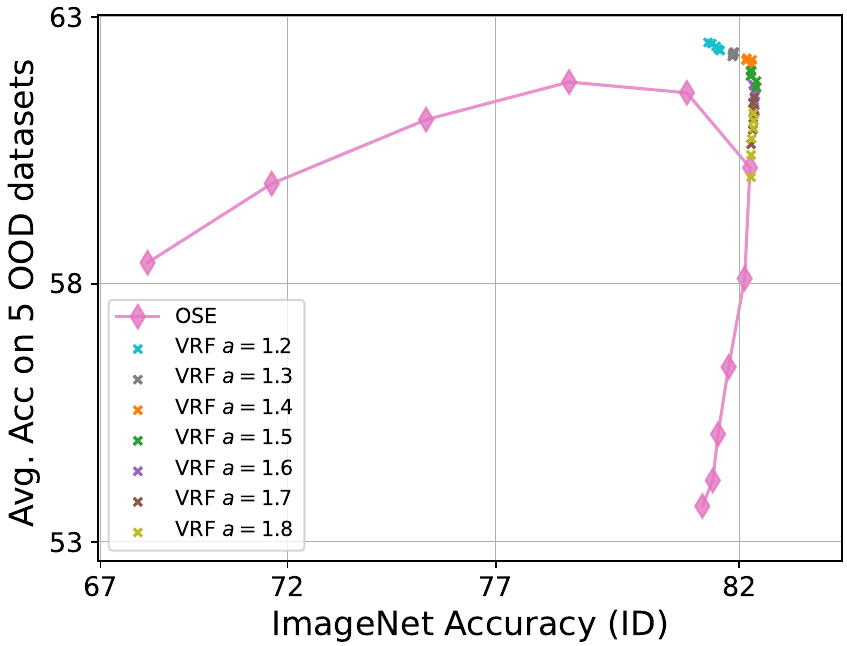}
        \caption{ID-OOD scatters for VRF on ImageNet and its variants.}
        \label{fig:scatter_borderrange}
    \end{minipage}
    \hfill
    \begin{minipage}[t]{0.32\textwidth}
        \centering
\includegraphics[width=\linewidth]{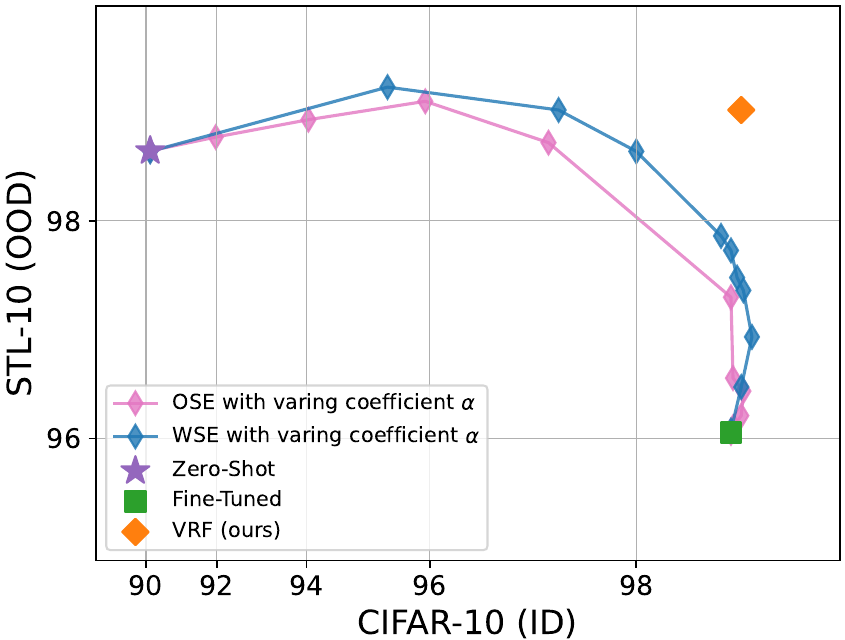}
        \caption{ID-OOD frontier curves for OSE and WSE for CIFAR-10 $\rightarrow$ STL10.}
        \label{fig:cifar10_wse}
    \end{minipage}%
    \hfill
    \begin{minipage}[t]{0.32\textwidth}
        \centering
\includegraphics[width=\linewidth]{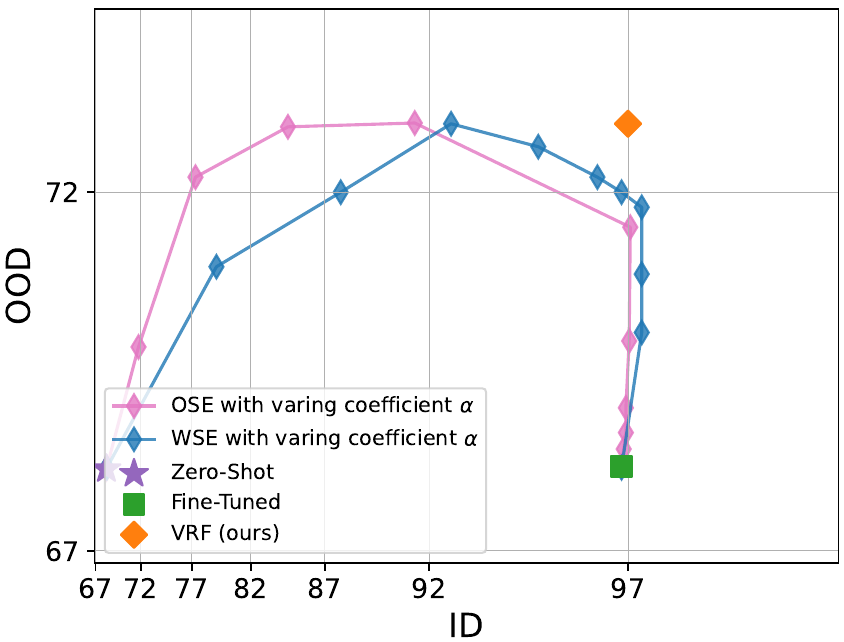}
        \caption{ID-OOD frontier curves for OSE and WSE for Entity-30.}
        \label{fig:entity_wse}
    \end{minipage}%    
\end{figure}

\subsection{Plotting ID-OOD Trade-Offs of VRF}\label{sec:vrf_tradeoffs}
In Figure~\ref{fig:scatter_borderrange}, we present an ID-OOD scatter plot over a wide range of $a$ and $b$ values on ImageNet and its five variants using CLIP ViT-B/16. Specifically, $a$ varies from 1.2 to 1.8, while $b$ ranges from 0.5 to 1.0.
Our method consistently achieves better ID-OOD trade-offs, as indicated by its points lying outside the OSE curve (represented by the magenta curve) across different configurations.

\begin{table}[t]
 \caption{Accuracy of E2E-FT based OSE on ImageNet and derived distribution shifts for various values of the mixing coefficient $\alpha$. Results shown for CLIP ViT-B/16.}
 \label{tab:mixing_alpha}
  \centering
    \tabstyle{6pt}
    \begin{tabular}{l c|ccccc|c}
    \toprule
      \multirow{2}{*}{$\alpha$} & \multirow{2}{*}{IN} & \multicolumn{5}{c|}{Distribution shifts} & Avg \\  
     &  & IN-V2 & IN-Sketch & IN-A & IN-R & ObjectNet & shifts \\
    \midrule
    0.0 &  68.3 & 61.9 & 48.3 & 50.1 &	77.6 &54.2 & 58.4\\
    0.1 &71.6	&64.8 &	49.9 &	50.7	&78.5&	55.5 &	59.9 \\
    0.2 & 75.4	& 67.5	& 51.6	& 51.2	& 79.2	& 56.3	& 61.1 \\
    0.3	&78.6	&69.7	&52.3	&50.9	&79.3	&56.8	&{61.8} \\
    0.4	&81.0   &71.3&52.0 &	49.6 &	78.6	& 56.6	& 61.6 \\
    0.5 &{82.2} & 72.0 &	50.6 &	46.8 &	76.7 & 54.9 & 60.2 \\
    0.6	&82.1 & 71.6	& 48.7 &	42.9& 	73.8 & 	53.3 &	58.1 \\
    0.7 &	81.8 &	71.2 &	47.3 &	40.5	& 71.1 &	52.2	& 56.4 \\
    0.8	&81.6	&70.9	&46.3	&38.5	&68.3	&51.4	&55.1 \\
    0.9	& 81.5	& 70.7	& 45.6	& 37.4	& 66.6	& 50.8	& 54.2 \\
    1.0 & 81.3 & 70.6 & 45.1 & 36.6 & 65.6 & 50.5 & 53.7\\ \bottomrule
    \end{tabular}
\end{table}

\begin{table}[t]
 \caption{Accuracy of E2E-FT based WSE on ImageNet and derived distribution shifts for various values of the mixing coefficient $\alpha$. Results shown for CLIP ViT-B/16.}
 \label{tab:mixing_alpha_wse}
  \centering
    \tabstyle{6pt}
    \begin{tabular}{l c|ccccc|c}
    \toprule
      \multirow{2}{*}{$\alpha$} & \multirow{2}{*}{IN} & \multicolumn{5}{c|}{Distribution shifts} & Avg \\  
     &  & IN-V2 & IN-Sketch & IN-A & IN-R & ObjectNet & shifts \\
    \midrule
    0.0 &  68.3 & 61.9 & 48.3 & 50.1 &	77.6 &54.2 & 58.4\\
    0.1 & 72.9 & 65.7 &	50.8 & 52.5		& 79.4 & 55.7 & 60.8	 \\
    0.2 & 76.4	& 68.7	& 52.5	& 54.2	& 80.1	& 57.1	& 62.5 \\
    0.3	& 78.9	& 70.6	& 53.6	&	54.6 & 80.1	& 57.5	& 63.3 \\
    0.4	& 80.5  & 72.1 & 54.1& 53.8	 &79.6	& 57.7	& 63.5\\
    0.5 & 81.7 & 72.8 &	53.9 & 52.2 &	78.7 &  57.3& 63.0 \\
    0.6	& 82.4& 72.9 & 53.4 &50.0	& 77.2 & 56.2 &	 61.9\\
    0.7 & 82.5 & 73.2 &	52.4 &47.4	& 75.2 &55.0	& 60.6\\
    0.8	& 82.5	& 72.8	& 51.0	& 44.6	&72.7	&	53.5&58.9 \\
    0.9	& 	82.1 & 72.0	& 48.9	&	40.9& 69.5	& 51.7	& 56.6 \\
    1.0 & 81.3 & 70.6 & 45.1 & 36.6 & 65.6 & 50.5 & 53.7\\ \bottomrule
    \end{tabular}
\end{table}
\begin{table}
 \caption{Accuracy of E2E-FT based VRF on ImageNet and derived distribution shifts for various values of $a$ and $b$. Results shown for CLIP ViT-B/16.}
 \label{tab:mixing_ab}
  \centering
    \tabstyle{6pt}
    \begin{tabular}{ll c|ccccc|c}
    \toprule
      \multirow{2}{*}{$a$} & \multirow{2}{*}{$b$} & \multirow{2}{*}{IN} & \multicolumn{5}{c|}{Distribution shifts} & Avg \\  
    & &  & IN-V2 & IN-Sketch & IN-A & IN-R & ObjectNet & shifts \\
    \midrule
    1.4 & 0.5 & 82.2  & 72.2 & 52.7 &  49.6  &  79.4 & 56.7 & 62.1 \\ 
    1.4 & 0.6 & 82.2 & 72.2 &  52.7  &  49.6 &  79.4 &  56.7 &  62.1 \\ 
    1.4 & 0.7 & 82.2 & 72.2 & 52.7 & 49.7 & 79.4 & 56.8 & 62.2 \\ 
    1.4 & 0.8 & 82.1 & 72.2 & 52.7 & 49.7 & 79.4 & 56.8 & 62.2\\ 
    1.4 & 0.9 & 82.1 & 72.1 & 52.7 & 49.7 & 79.4 &56.8 & 62.2 \\ 
    1.4 & 1.0 & 82.1 & 72.1 & 52.7 & 49.7 & 79.4 & 56.8 & 62.2 \\ \midrule
    1.5 & 0.5 & 82.3 & 72.1 & 52.3 & 48.7 &79.0& 56.2 & 61.7 \\
    1.5 & 0.6 & 82.3 & 72.1 & 52.4 & 48.9 & 79.1 & 56.4 & 61.8\\
    1.5 & 0.7 & 82.2 & 72.2 & 52.4 & 49.1 & 79.2 & 56.5 & 61.9 \\
    1.5 & 0.8 & 82.2 & 72.2 & 52.5 & 49.2 & 79.2 & 56.6 & 61.9 \\
    1.5 & 0.9 & 82.2 & 72.2 & 52.5 & 49.3 & 79.2 & 56.6 & 62.0 \\
    1.5 & 1.0 & 82.2 & 72.2 & 52.6 & 49.4 & 79.2 & 56.6 & 62.0 \\ \midrule
    1.6 & 0.5 & 82.3 & 71.9 & 51.9 & 48.0 & 78.5 & 55.7 & 61.2 \\ 
    1.6 & 0.6 & 82.3 & 72.1 & 52.1 & 48.3 & 78.6 & 55.9 & 61.4 \\
    1.6 & 0.7 & 82.3 & 72.1 & 52.2 & 48.5 & 78.8 & 56.0 & 61.5 \\
    1.6 & 0.8 & 82.3 & 72.2 & 52.3 & 48.6 & 78.9 & 56.1 & 61.6 \\
    1.6 & 0.9 & 82.3 & 72.2 & 52.3 & 48.7 & 79.0 & 56.2 & 61.7 \\
    1.6 & 1.0 & 82.3 & 72.1 & 52.4 & 48.8 & 79.0 & 56.3 & 61.7 \\
    \bottomrule
    \end{tabular}
\end{table}
\subsection{Additional Results for OSE, WSE and VRF with Varying Hyper-Parameters}

Results for all mixing coefficient $\alpha$ for OSE and WSE are available in Table~\ref{tab:mixing_alpha} and Table~\ref{tab:mixing_alpha_wse}, respectively. 
Results for values of $a$ and $b$ are available in Table~\ref{tab:mixing_ab}. In addition, we plot the ID-OOD trade-off curves for OSE and WSE on the CIFAR-10 and Entity-30 datasets in Figures~\ref{fig:cifar10_wse} and~\ref{fig:entity_wse}, respectively.

\begin{table}
\caption{Optimal Results search on test set on ImageNet and its five variants for CLIP ViT-B/16.}
\label{tab:oracle}
  \centering
    \tabstyle{4pt}
    \begin{tabular}{l c|ccccc|c}
    \toprule
      \multirow{2}{*}{Method} & \multirow{2}{*}{IN} & \multicolumn{5}{c|}{Distribution shifts} & Avg \\  
     &  & IN-V2 & IN-Sketch & IN-A & IN-R & ObjectNet & shifts \\
    \midrule
    E2E-FT &  81.3 &	70.6 &45.1 	 & 36.6 &65.6 &50.5  & 53.7  \\
    \quad + \Ours (ours) &  82.3& 72.1 & 52.9  & 48.4   & 78.7 &56.4  & 61.8 \\
    \quad + \Ours (oracle) & 82.3 & 72.2& 53.0 & 51.4& 79.7 & 57.9 & 62.9 \\
    \bottomrule
    \end{tabular}
\end{table}

\begin{table}
\caption{Applying VRF to other robust fine-tuning methods.}
\label{tab:flyp}
  \centering
    \tabstyle{4pt}
    \begin{tabular}{l c|ccccc|c}
    \toprule
      \multirow{2}{*}{Method} & \multirow{2}{*}{IN} & \multicolumn{5}{c|}{Distribution shifts} & Avg \\  
     &  & IN-V2 & IN-Sketch & IN-A & IN-R & ObjectNet & shifts \\
    \midrule
    FLYP~\cite{goyal2022finetune} &  82.6&	73.0 	&71.4&	48.1&	49.6&	58.7&	60.2 \\
    \quad + WSE &  82.9 &	73.5&	76.0&	53.0	&52.3&	60.8	&63.1 \\
    \quad + OSE & 82.8  &	73.6 &	77.0&	52.5	&51.9	&59.9	&62.8 \\
    \quad + \Ours & 82.8&	73.6	&78.6&	52.9&	53.0	&61.2&	64.0 \\

    \bottomrule
    \end{tabular}
\end{table}
\subsection{Optimal Performance Searched on Test Sets}

We have conducted additional experiments where we optimized the hyperparameters for each test set of the ImageNet benckmarks. The results are summarized in  Table~\ref{tab:oracle}. 

\subsection{Combining VRF with Other Robust Fine-Tuning Methods}

Our VRF framework is orthogonal and complementary to existing fine-tuned models. To demonstrate this, we integrated FLYP~\cite{goyal2022finetune} into our VRF framework. The results in Table~\ref{tab:flyp} show that VRF improves OSE's performance under distribution shift by 1.1\% without compromising in-distribution (ID) performance.

\subsection{Additional Results for Selective Prediction Using OOD Detectors}\label{app:sp}

\begin{table}
 \caption{Breakdown performance for selective prediction using OOD detector.}
 \label{tab:addition_sp}
  \centering
    \tabstyle{6pt}
    \begin{tabular}{l c|ccccc|c}
    \toprule
      \multirow{2}{*}{OOD Detector} & \multirow{2}{*}{IN} & \multicolumn{5}{c|}{Distribution shifts} & Avg \\  
     &  & IN-V2 & IN-Sketch & IN-A & IN-R & ObjectNet & shifts \\ \midrule
     MSP~\cite{hendrycks2016baseline}     & 81.5 & 71.1 & 48.6 & 42.1 & 71.6 & 52.9 & 57.3\\
    Energy~\cite{liu2020energy}  &  81.0 &  70.5 &  48.1 &	42.3 &	73.9 &	53.0 & 57.6 \\
    MD~\cite{lee2018simple}      & 81.0 &70.4 &	49.7	& 41.7&	74.1&	52.6 & 57.7 \\
    kNN~\cite{sun2022out}     & 80.8 & 70.4 &	49.5	 &43.6 &	74.8 &	53.6 &  58.4  \\
    RMD~\cite{ren2021simple}     & 81.1 &70.6	&49.6	&44.4&	74.4&	53.1 & 58.4 \\
     \bottomrule
    \end{tabular}
\end{table}
We provide the breakdown performance for selective prediction using OOD detectors in Table~\ref{tab:addition_sp}.

\subsection{Curves of $\frac{\text{Acc}_\mathsf{ft}}{\text{Acc}_\mathsf{zs}}$ for ImageNet and its Five Distribution Shifted Datasets}\label{app:ft_zs_ratio}
\begin{figure}
    \centering
\includegraphics[width=0.6\textwidth]{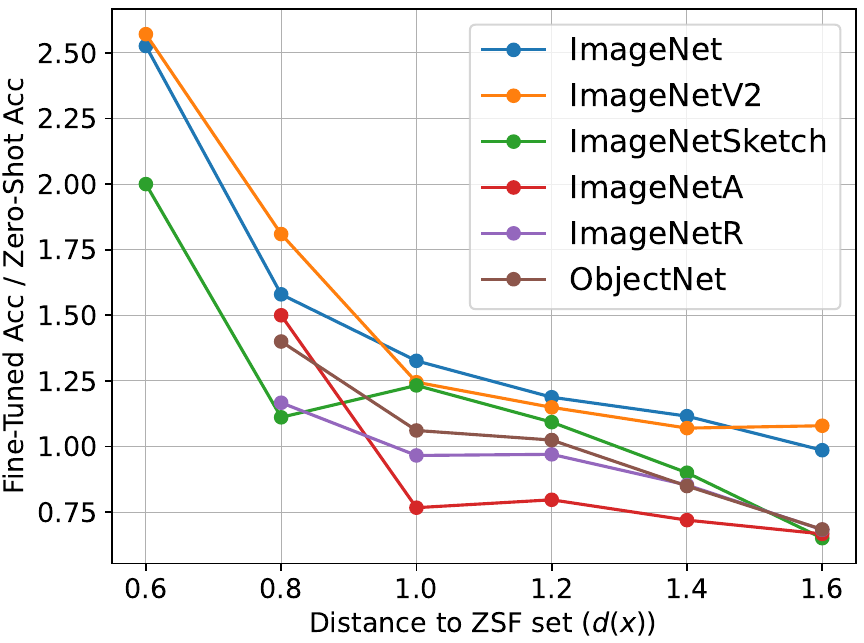}
    \caption{Relationship between $\frac{\text{Acc}_\mathsf{ft}}{\text{Acc}_\mathsf{zs}}$ and $d(\x)$ on ImageNet benchmarks.}
    \label{fig:ft_zs_all_datasets}
\end{figure}
In Figure~\ref{fig:ft_zs_all_datasets}, we examine the relationship between $\frac{\text{Acc}_\mathsf{ft}}{\text{Acc}_\mathsf{zs}}$ and $d(\x)$ for ImageNet and its five derived distribution shifted datasets. 
Based on the value of $d(\x)$, test samples are grouped into bins, and we compute
the ratio of fine-tuned accuracy to zero-shot accuracy for each bin.
For example, to compute the value of $\frac{\text{Acc}_\mathsf{ft}}{\text{Acc}_\mathsf{zs}}$ at $d(\x)=0.8$, we first identify the samples with $d(\x)\in [0.7, 0.9]$, then compute the averaged accuracy for these samples using zero-shot models and fine-tuned models, and finally compute the ratio.
Note that the averaged ratio $\frac{\text{Acc}_\mathsf{ft}}{\text{Acc}_\mathsf{zs}}$ on ImageNet-\{A,R\} and ObjectNet is undefined for $d(\x)=0.6$. This is because in these datasets, the zero-shot accuracy around $d(\x)=0.6$ is 0.
We observe that the trend of the ratio $\frac{\text{Acc}_\mathsf{ft}}{\text{Acc}_\mathsf{zs}}$ decreasing as $d(\x)$ increasing is stable for all ImageNet related datasets.

\clearpage

\section*{NeurIPS Paper Checklist}

\begin{enumerate}

\item {\bf Claims}
    \item[] Question: Do the main claims made in the abstract and introduction accurately reflect the paper's contributions and scope?
    \item[] Answer: \answerYes{} % Replace by \answerYes{}, \answerNo{}, or \answerNA{}.
    \item[] Justification: We point out that the ID-OOD trade-offs still exist in existing fine-tuning methods, and propose \Ours to simultaneously attain the best ID and OOD accuracy. Experiments on a variety of different models and tasks validate the effectiveness of our proposed method.
    \item[] Guidelines:
    \begin{itemize}
        \item The answer NA means that the abstract and introduction do not include the claims made in the paper.
        \item The abstract and/or introduction should clearly state the claims made, including the contributions made in the paper and important assumptions and limitations. A No or NA answer to this question will not be perceived well by the reviewers. 
        \item The claims made should match theoretical and experimental results, and reflect how much the results can be expected to generalize to other settings. 
        \item It is fine to include aspirational goals as motivation as long as it is clear that these goals are not attained by the paper. 
    \end{itemize}

\item {\bf Limitations}
    \item[] Question: Does the paper discuss the limitations of the work performed by the authors?
    \item[] Answer: \answerYes{} % Replace by \answerYes{}, \answerNo{}, or \answerNA{}.
    \item[] Justification: The limitations have been discussed in Section~\ref{sec:conclusion}.
    \item[] Guidelines:
    \begin{itemize}
        \item The answer NA means that the paper has no limitation while the answer No means that the paper has limitations, but those are not discussed in the paper. 
        \item The authors are encouraged to create a separate "Limitations" section in their paper.
        \item The paper should point out any strong assumptions and how robust the results are to violations of these assumptions (e.g., independence assumptions, noiseless settings, model well-specification, asymptotic approximations only holding locally). The authors should reflect on how these assumptions might be violated in practice and what the implications would be.
        \item The authors should reflect on the scope of the claims made, e.g., if the approach was only tested on a few datasets or with a few runs. In general, empirical results often depend on implicit assumptions, which should be articulated.
        \item The authors should reflect on the factors that influence the performance of the approach. For example, a facial recognition algorithm may perform poorly when image resolution is low or images are taken in low lighting. Or a speech-to-text system might not be used reliably to provide closed captions for online lectures because it fails to handle technical jargon.
        \item The authors should discuss the computational efficiency of the proposed algorithms and how they scale with dataset size.
        \item If applicable, the authors should discuss possible limitations of their approach to address problems of privacy and fairness.
        \item While the authors might fear that complete honesty about limitations might be used by reviewers as grounds for rejection, a worse outcome might be that reviewers discover limitations that aren't acknowledged in the paper. The authors should use their best judgment and recognize that individual actions in favor of transparency play an important role in developing norms that preserve the integrity of the community. Reviewers will be specifically instructed to not penalize honesty concerning limitations.
    \end{itemize}

\item {\bf Theory Assumptions and Proofs}
    \item[] Question: For each theoretical result, does the paper provide the full set of assumptions and a complete (and correct) proof?
    \item[] Answer: \answerYes{} % Replace by \answerYes{}, \answerNo{}, or \answerNA{}.
    \item[] Justification: In Section~\ref{sec:justify}, we prove that our \Ours can effectively reduce the variance of the ensemble model and thus achieve lower errors.
    \item[] Guidelines:
    \begin{itemize}
        \item The answer NA means that the paper does not include theoretical results. 
        \item All the theorems, formulas, and proofs in the paper should be numbered and cross-referenced.
        \item All assumptions should be clearly stated or referenced in the statement of any theorems.
        \item The proofs can either appear in the main paper or the supplemental material, but if they appear in the supplemental material, the authors are encouraged to provide a short proof sketch to provide intuition. 
        \item Inversely, any informal proof provided in the core of the paper should be complemented by formal proofs provided in appendix or supplemental material.
        \item Theorems and Lemmas that the proof relies upon should be properly referenced. 
    \end{itemize}

    \item {\bf Experimental Result Reproducibility}
    \item[] Question: Does the paper fully disclose all the information needed to reproduce the main experimental results of the paper to the extent that it affects the main claims and/or conclusions of the paper (regardless of whether the code and data are provided or not)?
    \item[] Answer: \answerYes{} % Replace by \answerYes{}, \answerNo{}, or \answerNA{}.
    \item[] Justification: We have provided the \Ours algorithm in Algorithm~\ref{algo:1} with descriptions in Section~\ref{sec:methods}, and included the implementation details in Section~\ref{sec:exp} and Section~\ref{app:exp-detail}.
    \item[] Guidelines:
    \begin{itemize}
        \item The answer NA means that the paper does not include experiments.
        \item If the paper includes experiments, a No answer to this question will not be perceived well by the reviewers: Making the paper reproducible is important, regardless of whether the code and data are provided or not.
        \item If the contribution is a dataset and/or model, the authors should describe the steps taken to make their results reproducible or verifiable. 
        \item Depending on the contribution, reproducibility can be accomplished in various ways. For example, if the contribution is a novel architecture, describing the architecture fully might suffice, or if the contribution is a specific model and empirical evaluation, it may be necessary to either make it possible for others to replicate the model with the same dataset, or provide access to the model. In general. releasing code and data is often one good way to accomplish this, but reproducibility can also be provided via detailed instructions for how to replicate the results, access to a hosted model (e.g., in the case of a large language model), releasing of a model checkpoint, or other means that are appropriate to the research performed.
        \item While NeurIPS does not require releasing code, the conference does require all submissions to provide some reasonable avenue for reproducibility, which may depend on the nature of the contribution. For example
        \begin{enumerate}
            \item If the contribution is primarily a new algorithm, the paper should make it clear how to reproduce that algorithm.
            \item If the contribution is primarily a new model architecture, the paper should describe the architecture clearly and fully.
            \item If the contribution is a new model (e.g., a large language model), then there should either be a way to access this model for reproducing the results or a way to reproduce the model (e.g., with an open-source dataset or instructions for how to construct the dataset).
            \item We recognize that reproducibility may be tricky in some cases, in which case authors are welcome to describe the particular way they provide for reproducibility. In the case of closed-source models, it may be that access to the model is limited in some way (e.g., to registered users), but it should be possible for other researchers to have some path to reproducing or verifying the results.
        \end{enumerate}
    \end{itemize}

\item {\bf Open access to data and code}
    \item[] Question: Does the paper provide open access to the data and code, with sufficient instructions to faithfully reproduce the main experimental results, as described in supplemental material?
    \item[] Answer: \answerYes{} % Replace by \answerYes{}, \answerNo{}, or \answerNA{}.
    \item[] Justification: We have uploaded the codes in supplemental material.
    \item[] Guidelines:
    \begin{itemize}
        \item The answer NA means that paper does not include experiments requiring code.
        \item Please see the NeurIPS code and data submission guidelines (\url{https://nips.cc/public/guides/CodeSubmissionPolicy}) for more details.
        \item While we encourage the release of code and data, we understand that this might not be possible, so “No” is an acceptable answer. Papers cannot be rejected simply for not including code, unless this is central to the contribution (e.g., for a new open-source benchmark).
        \item The instructions should contain the exact command and environment needed to run to reproduce the results. See the NeurIPS code and data submission guidelines (\url{https://nips.cc/public/guides/CodeSubmissionPolicy}) for more details.
        \item The authors should provide instructions on data access and preparation, including how to access the raw data, preprocessed data, intermediate data, and generated data, etc.
        \item The authors should provide scripts to reproduce all experimental results for the new proposed method and baselines. If only a subset of experiments are reproducible, they should state which ones are omitted from the script and why.
        \item At submission time, to preserve anonymity, the authors should release anonymized versions (if applicable).
        \item Providing as much information as possible in supplemental material (appended to the paper) is recommended, but including URLs to data and code is permitted.
    \end{itemize}

\item {\bf Experimental Setting/Details}
    \item[] Question: Does the paper specify all the training and test details (e.g., data splits, hyperparameters, how they were chosen, type of optimizer, etc.) necessary to understand the results?
    \item[] Answer: \answerYes{} % Replace by \answerYes{}, \answerNo{}, or \answerNA{}.
    \item[] Justification:We have provided all the training and testing details in Section~\ref{sec:exp} and Section~\ref{app:exp-detail}.
    \item[] Guidelines:
    \begin{itemize}
        \item The answer NA means that the paper does not include experiments.
        \item The experimental setting should be presented in the core of the paper to a level of detail that is necessary to appreciate the results and make sense of them.
        \item The full details can be provided either with the code, in appendix, or as supplemental material.
    \end{itemize}

\item {\bf Experiment Statistical Significance}
    \item[] Question: Does the paper report error bars suitably and correctly defined or other appropriate information about the statistical significance of the experiments?
    \item[] Answer: \answerNo{} % Replace by \answerYes{}, \answerNo{}, or \answerNA{}.
    \item[] Justification: Given the zero-shot and the fine-tuned models, the process of our post-hoc method is deterministic. Run multiple times will not introduce randomness.
    \item[] Guidelines:
    \begin{itemize}
        \item The answer NA means that the paper does not include experiments.
        \item The authors should answer "Yes" if the results are accompanied by error bars, confidence intervals, or statistical significance tests, at least for the experiments that support the main claims of the paper.
        \item The factors of variability that the error bars are capturing should be clearly stated (for example, train/test split, initialization, random drawing of some parameter, or overall run with given experimental conditions).
        \item The method for calculating the error bars should be explained (closed form formula, call to a library function, bootstrap, etc.)
        \item The assumptions made should be given (e.g., Normally distributed errors).
        \item It should be clear whether the error bar is the standard deviation or the standard error of the mean.
        \item It is OK to report 1-sigma error bars, but one should state it. The authors should preferably report a 2-sigma error bar than state that they have a 96\% CI, if the hypothesis of Normality of errors is not verified.
        \item For asymmetric distributions, the authors should be careful not to show in tables or figures symmetric error bars that would yield results that are out of range (e.g. negative error rates).
        \item If error bars are reported in tables or plots, The authors should explain in the text how they were calculated and reference the corresponding figures or tables in the text.
    \end{itemize}

\item {\bf Experiments Compute Resources}
    \item[] Question: For each experiment, does the paper provide sufficient information on the computer resources (type of compute workers, memory, time of execution) needed to reproduce the experiments?
    \item[] Answer: \answerYes{} % Replace by \answerYes{}, \answerNo{}, or \answerNA{}.
    \item[] Justification: We have provided the GPU type and number to reproduce the results in Section~\ref{app:exp-detail}.
    \item[] Guidelines:
    \begin{itemize}
        \item The answer NA means that the paper does not include experiments.
        \item The paper should indicate the type of compute workers CPU or GPU, internal cluster, or cloud provider, including relevant memory and storage.
        \item The paper should provide the amount of compute required for each of the individual experimental runs as well as estimate the total compute. 
        \item The paper should disclose whether the full research project required more compute than the experiments reported in the paper (e.g., preliminary or failed experiments that didn't make it into the paper). 
    \end{itemize}
    
\item {\bf Code Of Ethics}
    \item[] Question: Does the research conducted in the paper conform, in every respect, with the NeurIPS Code of Ethics \url{https://neurips.cc/public/EthicsGuidelines}?
    \item[] Answer: \answerYes{} % Replace by \answerYes{}, \answerNo{}, or \answerNA{}.
    \item[] Justification: We have reviewed the NeurIPS Code of Ethics.
    \item[] Guidelines:
    \begin{itemize}
        \item The answer NA means that the authors have not reviewed the NeurIPS Code of Ethics.
        \item If the authors answer No, they should explain the special circumstances that require a deviation from the Code of Ethics.
        \item The authors should make sure to preserve anonymity (e.g., if there is a special consideration due to laws or regulations in their jurisdiction).
    \end{itemize}

\item {\bf Broader Impacts}
    \item[] Question: Does the paper discuss both potential positive societal impacts and negative societal impacts of the work performed?
    \item[] Answer: \answerYes{} % Replace by \answerYes{}, \answerNo{}, or \answerNA{}.
    \item[] Justification: We have discussed the societal impacts in Section~\ref{sec:conclusion}.
    \item[] Guidelines:
    \begin{itemize}
        \item The answer NA means that there is no societal impact of the work performed.
        \item If the authors answer NA or No, they should explain why their work has no societal impact or why the paper does not address societal impact.
        \item Examples of negative societal impacts include potential malicious or unintended uses (e.g., disinformation, generating fake profiles, surveillance), fairness considerations (e.g., deployment of technologies that could make decisions that unfairly impact specific groups), privacy considerations, and security considerations.
        \item The conference expects that many papers will be foundational research and not tied to particular applications, let alone deployments. However, if there is a direct path to any negative applications, the authors should point it out. For example, it is legitimate to point out that an improvement in the quality of generative models could be used to generate deepfakes for disinformation. On the other hand, it is not needed to point out that a generic algorithm for optimizing neural networks could enable people to train models that generate Deepfakes faster.
        \item The authors should consider possible harms that could arise when the technology is being used as intended and functioning correctly, harms that could arise when the technology is being used as intended but gives incorrect results, and harms following from (intentional or unintentional) misuse of the technology.
        \item If there are negative societal impacts, the authors could also discuss possible mitigation strategies (e.g., gated release of models, providing defenses in addition to attacks, mechanisms for monitoring misuse, mechanisms to monitor how a system learns from feedback over time, improving the efficiency and accessibility of ML).
    \end{itemize}
    
\item {\bf Safeguards}
    \item[] Question: Does the paper describe safeguards that have been put in place for responsible release of data or models that have a high risk for misuse (e.g., pretrained language models, image generators, or scraped datasets)?
    \item[] Answer: \answerNA{} % Replace by \answerYes{}, \answerNo{}, or \answerNA{}.
    \item[] Justification: Our method aims to improve the robustness when fine-tuning models, which poses no such risks to the best of our knowledge.
    \item[] Guidelines:
    \begin{itemize}
        \item The answer NA means that the paper poses no such risks.
        \item Released models that have a high risk for misuse or dual-use should be released with necessary safeguards to allow for controlled use of the model, for example by requiring that users adhere to usage guidelines or restrictions to access the model or implementing safety filters. 
        \item Datasets that have been scraped from the Internet could pose safety risks. The authors should describe how they avoided releasing unsafe images.
        \item We recognize that providing effective safeguards is challenging, and many papers do not require this, but we encourage authors to take this into account and make a best faith effort.
    \end{itemize}

\item {\bf Licenses for existing assets}
    \item[] Question: Are the creators or original owners of assets (e.g., code, data, models), used in the paper, properly credited and are the license and terms of use explicitly mentioned and properly respected?
    \item[] Answer: \answerYes{} % Replace by \answerYes{}, \answerNo{}, or \answerNA{}.
    \item[] Justification: We have provided the licences of each dataset in Section~\ref{app:licenses}.
    \item[] Guidelines:
    \begin{itemize}
        \item The answer NA means that the paper does not use existing assets.
        \item The authors should cite the original paper that produced the code package or dataset.
        \item The authors should state which version of the asset is used and, if possible, include a URL.
        \item The name of the license (e.g., CC-BY 4.0) should be included for each asset.
        \item For scraped data from a particular source (e.g., website), the copyright and terms of service of that source should be provided.
        \item If assets are released, the license, copyright information, and terms of use in the package should be provided. For popular datasets, \url{paperswithcode.com/datasets} has curated licenses for some datasets. Their licensing guide can help determine the license of a dataset.
        \item For existing datasets that are re-packaged, both the original license and the license of the derived asset (if it has changed) should be provided.
        \item If this information is not available online, the authors are encouraged to reach out to the asset's creators.
    \end{itemize}

\item {\bf New Assets}
    \item[] Question: Are new assets introduced in the paper well documented and is the documentation provided alongside the assets?
    \item[] Answer: \answerNA{} % Replace by \answerYes{}, \answerNo{}, or \answerNA{}.
    \item[] Justification: We do not release new assets.
    \item[] Guidelines:
    \begin{itemize}
        \item The answer NA means that the paper does not release new assets.
        \item Researchers should communicate the details of the dataset/code/model as part of their submissions via structured templates. This includes details about training, license, limitations, etc. 
        \item The paper should discuss whether and how consent was obtained from people whose asset is used.
        \item At submission time, remember to anonymize your assets (if applicable). You can either create an anonymized URL or include an anonymized zip file.
    \end{itemize}

\item {\bf Crowdsourcing and Research with Human Subjects}
    \item[] Question: For crowdsourcing experiments and research with human subjects, does the paper include the full text of instructions given to participants and screenshots, if applicable, as well as details about compensation (if any)? 
    \item[] Answer: \answerNA{} % Replace by \answerYes{}, \answerNo{}, or \answerNA{}.
    \item[] Justification: We do not involve such experiments.
    \item[] Guidelines:
    \begin{itemize}
        \item The answer NA means that the paper does not involve crowdsourcing nor research with human subjects.
        \item Including this information in the supplemental material is fine, but if the main contribution of the paper involves human subjects, then as much detail as possible should be included in the main paper. 
        \item According to the NeurIPS Code of Ethics, workers involved in data collection, curation, or other labor should be paid at least the minimum wage in the country of the data collector. 
    \end{itemize}

\item {\bf Institutional Review Board (IRB) Approvals or Equivalent for Research with Human Subjects}
    \item[] Question: Does the paper describe potential risks incurred by study participants, whether such risks were disclosed to the subjects, and whether Institutional Review Board (IRB) approvals (or an equivalent approval/review based on the requirements of your country or institution) were obtained?
    \item[] Answer: \answerNA{} % Replace by \answerYes{}, \answerNo{}, or \answerNA{}.
    \item[] Justification: This paper does not involve crowdsourcing nor research with human subjects.
    \item[] Guidelines:
    \begin{itemize}
        \item The answer NA means that the paper does not involve crowdsourcing nor research with human subjects.
        \item Depending on the country in which research is conducted, IRB approval (or equivalent) may be required for any human subjects research. If you obtained IRB approval, you should clearly state this in the paper. 
        \item We recognize that the procedures for this may vary significantly between institutions and locations, and we expect authors to adhere to the NeurIPS Code of Ethics and the guidelines for their institution. 
        \item For initial submissions, do not include any information that would break anonymity (if applicable), such as the institution conducting the review.
    \end{itemize}

\end{enumerate}

\end{document}